\documentclass[preprint, review ,12pt]{elsarticle}

\usepackage{amsmath}
\usepackage{algorithm}
\usepackage{algpseudocode}
\usepackage{amsfonts}
\usepackage{amssymb}
\usepackage{lmodern}
\usepackage{graphicx}
\usepackage{xcolor}
\usepackage{multirow}
\usepackage{tikz}
\usepackage{lipsum}
\usepackage{cancel}
\usepackage{bm}         
\usepackage{mathtools, nccmath}
\usepackage{xspace}
\usepackage{bbm}
\usepackage{siunitx} 
\usepackage[normalem]{ulem}
\usepackage{comment}
\usepackage{booktabs}	
\usepackage{float}	
\usepackage{subcaption} 
\usepackage{xr}
\usepackage{soul}

\usepackage[margin=0.8in]{geometry} 

\newcommand{\matr}[1]{\mathbf{{#1}}}    
\newcommand{\ve}[1]{\bm{{#1}}}          
\newcommand{\x}{\ve{x}}

\newcommand{\vtheta}{\ve{\theta}}
\newcommand{\zeros}{\matr{0}}
\newcommand{\eyei}{\matr{I}}
\newcommand{\ones}{\ve{1}}
\newcommand{\brc}[1]{\left(#1\right)}  
\newcommand{\cbk}[1]{\left\{#1\right\}}     
\newcommand{\sbk}[1]{\left[#1\right]}       

\newcommand{\pr}[1]{p\brc{#1}}
\usepackage{caption}
\usepackage{subcaption}
\newcommand{\mSp}{\matr{S}_p} 
\newcommand{\mSu}{\matr{S}_u} 

\usepackage{enumitem}
\newlist{steps}{enumerate}{1}
\setlist[steps, 1]{label = \textbf{Step} \arabic*:}

\usepackage{hyperref}
 \hypersetup{
 colorlinks,
 linktocpage,
     linkcolor=red,
     citecolor=red,
    urlcolor=red
}
\usepackage[capitalize]{cleveref}

\begin{document}

\begin{frontmatter}



\title{Probabilistic Digital Twin for Misspecified Structural Dynamical Systems via Latent Force Modeling and Bayesian Neural Networks}

\author[label1]{Sahil Kashyap\corref{cor1}}\ead{amz228088@am.iitd.ac.in}
\author[label1]{Rajdip Nayek} \ead{rajdipn@iitd.ac.in}

\cortext[cor1]{Corresponding author}

\affiliation[label1]{organization={Department of Applied Mechanics, Indian Institute of Technology Delhi},
            addressline={Hauz Khas}, 
            city={New Delhi},
            postcode={110016}, 
            state={Delhi},
            country={India}}

\begin{abstract}

This work presents a probabilistic digital twin framework for response prediction in dynamical systems governed by misspecified physics. The approach integrates Gaussian Process Latent Force Models (GPLFM) and Bayesian Neural Networks (BNNs) to enable end-to-end uncertainty-aware inference and prediction. In the diagnosis phase, model-form errors (MFEs) are treated as latent input forces to a nominal linear dynamical system and jointly estimated with system states using GPLFM from sensor measurements. A BNN is then trained on posterior samples to learn a probabilistic nonlinear mapping from system states to MFEs, while capturing diagnostic uncertainty. For prognosis, this mapping is used to generate pseudo-measurements, enabling state prediction via Kalman filtering. The framework allows for systematic propagation of uncertainty from diagnosis to prediction, a key capability for trustworthy digital twins. The framework is demonstrated using four nonlinear examples: a single degree of freedom (DOF) oscillator, a multi-DOF system, and two established benchmarks --- the Bouc-Wen hysteretic system and the Silverbox experimental dataset --- highlighting its predictive accuracy and robustness to model misspecification.
\end{abstract}



\end{frontmatter}

\section{Introduction} \label{sec:intro}
 
The analysis of dynamical systems underpins much of our understanding of physical processes. A fundamental objective in this context is the accurate prediction of system states, which enables key applications such as structural health monitoring (SHM) \cite{SHM1}, control \cite{control1}, decision-making \cite{decision1}, and design \cite{design1}. However, high-fidelity prediction in real-world systems is often impeded by model misspecification --- an inherent consequence of incomplete or simplified physics.

In recent years, the concept of digital twins has emerged as a powerful paradigm for enabling such predictive capabilities. Digital twins \cite{Review_digitalTwin2025,EJ_Cross_digital,S_mahadevan_DT} aim to create virtual counterparts of physical systems by integrating physics-based models, real-time sensor data, and data-driven methods. Their promise lies in enabling online monitoring, predictive control, and diagnostics. However, a major barrier to their reliability is the presence of model-form errors (MFEs) --- systematic discrepancies between modeled and actual system behavior.

Such discrepancies arise because governing equations of physical systems are rarely known exactly. Models are typically constructed using idealized assumptions for tractability, leading to gaps between the modeled and true dynamics. For instance, structural systems often exhibit local nonlinearities such as joint friction, nonlinear damping, or stiffness. Nevertheless, linear finite element models (FEM) remain standard in practice due to their simplicity and computational efficiency. While convenient, these models fail to capture essential nonlinear behavior. Even when nonlinear effects are anticipated, identifying their precise nature or location is often infeasible. Damping illustrates a similar issue: although its true physical mechanisms are complex, simplified forms such as Rayleigh damping are routinely used for mathematical convenience \cite{adhikari2006damping}. Collectively, such simplifications introduce MFEs \cite{2022RogersNonlinearIdentification,abhinav_zero_dynamics,kashyap2024gaussian}, which must be addressed for reliable state prediction.

Since MFEs are intrinsic to all practical models, accurate state prediction must begin with diagnosing these errors. In the structural dynamics context, MFEs can stem from unmodeled nonlinearities, uncertain parameters, imperfect boundary or initial conditions, or unknown external inputs. In this work, we assume that MFEs are state-dependent, i.e., they vary as functions of the instantaneous system states such as displacements and velocities. However, in practice, not all system states are states are measured; instead, only partial observations are typically available, usually limited to (noisy) acceleration measurements recorded by accelerometers. This makes it necessary not only to infer both the system states and the MFEs from noisy data, but also to establish the latent functional relationship that links the system states to the MFEs --- a relationship that is crucial for achieving accurate predictions.

The proposed framework aligns with the broader vision of digital twins by offering a probabilistic and interpretable approach to combine misspecified physics with noisy sensor data, and to predict system behavior under new test inputs.

While previous works often correct for MFEs by embedding a learned mapping directly into the governing equations, such corrected dynamics are usually propagated deterministically. This deterministic embedding disregards the uncertainty in the learned mapping and assumes stability under all unseen test inputs, an assumption that does not always hold in practice. More rigorous alternatives attempt probabilistic forward integration of the MFE-corrected system using Monte Carlo sampling; however, in our experience, such methods often lead to unstable trajectories due to the data-driven nature of the learned map, which lacks guarantees of dynamical stability. These limitations motivate the development of a probabilistic yet numerically stable approach for state prediction under misspecified physics. Our framework addresses this by using a GPLFM framework during prognosis, ensuring uncertainty-aware predictions without compromising stability.

We structure this task into three stages:
\begin{itemize}
\item \textbf{Diagnosis}: Estimate both the full system states and the latent MFEs using available sensor data and a partially specified physics-based model.

\item \textbf{Mapping}: Learn a functional relationship between the estimated system states and the inferred MFEs, assuming the latter are state-dependent.

\item \textbf{Prognosis}: Predict the system's future states under new excitations, without access to further measurements, while propagating uncertainty arising from the diagnosis stage.
\end{itemize}

Since the inference of MFEs and system states from noisy measurements is inherently uncertain, these uncertainties must be systematically accounted for during prognosis. This calls for a probabilistic framework capable of quantifying and propagating uncertainties throughout the pipeline.

\subsection{Related work}
 
Machine learning has increasingly been used to model dynamical systems from data, especially when governing equations are partially known or misspecified. Several works have applied deep neural networks (DNNs) to learn system behavior directly from measurements \cite{raissi_data_driven, RUDY2019483}. While effective in capturing complex dynamics, such purely data-driven models often require extensive training data under varied excitations and may fail to generalize to out-of-distribution inputs. Physics-based models, though more interpretable and generalizable, are rarely accurate in practice due to simplifying assumptions. This trade-off has motivated hybrid approaches that integrate partial physics with data-driven corrections, aiming to retain interpretability while improving predictive accuracy. These typically introduce an additive MFE term into the governing equations of the misspecified model, inferred using available sensor data \cite{kashyap2024gaussian,subramanian2023probabilistic,RileighBandy_MFE,Neal_MFE,GAN_MFE,takeishi2021physics,subramanian2022non,Rimple_Sandu}. 

Depending on the treatment of this additive term, hybrid approaches can be classified into white-box and black-box formulations. White-box methods aim for interpretability by representing MFEs as sparse combinations of basis functions drawn from a pre-defined library of candidate terms \cite{sindy,rudy2019data,champneys2024bindy,nayek2021spike,Rohan_paper} or maybe using grammar-based symbolic regression \cite{E_Chatzi_grammar,Grammar2}. However, these methods require careful library construction, which can be challenging in the absence of prior knowledge about the system's latent dynamics. 

Black-box methods bypass the need for interpretability by learning the MFE–state relationship directly from data. These include (but are not limited to) DNN-based models that approximate a deterministic mapping from system states to MFEs \cite{takeishi2021physics,blakseth2022deep}, as well as Gaussian process (GP) models that provide a probabilistic representation of the same \cite{nioche2024transformed,gardner2020bayesian}. GPs offer the advantage of uncertainty quantification in both the learned mapping and subsequent predictions.

In both paradigms, the learned MFE functions are reintroduced into the governing equations of the misspecified model to yield a corrected equation of motion (EoM), which can then be forward-solved using numerical integrators to yield predictions for unseen excitation.

An alternative class of approaches employs probabilistic Bayesian filtering and smoothing techniques \cite{sarkka2013bayesian} to jointly infer system states and MFEs. This paradigm, also adopted in the present work, enables simultaneous estimation of both quantities, facilitating the subsequent learning of a functional relationship between them. These methods typically augment the state-space model with an additional latent state representing the MFE, usually modeled as a random walk process. For instance, Subramanian et al.\ \cite{subramanian2019bayesian} and Garg et al.\ \cite{garg2022physics} modeled MFEs as random walks, estimating them via filtering and using ANNs \cite{subramanian2019bayesian} or GPs \cite{garg2022physics} to learn the functional map between states and MFEs. While these methods offer a principled probabilistic framework, the discrete-time random-walk assumption --- corresponding to a marginally stable continuous-time system --- can introduce drift in state estimates, particularly with acceleration-only measurements as shown in \cite{nayek2019gaussian}.

To address this drift, several works have adopted Gaussian Process Latent Force Models (GPLFM), which incorporate non-zero dynamics into the latent MFE process. These have been applied to structural systems in recent studies \cite{2022RogersNonlinearIdentification,kashyap2024gaussian,nayek2019gaussian,alice_cicirello_2022_GP,alice_2025_GPLFM,EM_lourens_GPLFM}. In particular, Kashyap et al. \cite{kashyap2024gaussian} applied GPLFM to an MDOF shear-storey structure with local nonlinearities. MFEs were modeled as stationary GPs and formulated in state-space form to facilitate joint inference with system states via Bayesian filtering. The GP hyperparameters were optimized offline, enabling high-fidelity state and MFE estimation using sensor data.

Across most prior work, prognosis is performed by forward solving the corrected EoM using numerical integrators, where the learned MFE–state mapping (usually the deterministic part) is directly embedded into the EoM. While intuitive, this approach can sometimes suffer from numerical instability when applied to (unseen) test inputs. Since the mapping is trained on finite data, there is no guarantee that it ensures system stability. The Monte Carlo-based forward integrations of this idea --- where one samples multiple realizations of the MFE from the learned distribution and propagates each realization using Monte Carlo integration --- face even more problems. The lack of dynamical constraints on the learned mapping often leads to divergent or unstable trajectories, in the experience of the authors. Thus, while Monte Carlo-based forward integration offers a theoretically sound means of uncertainty propagation, it is computationally expensive and numerically fragile. 

In contrast, the present work avoids direct embedding of the learned mapping into the system dynamics. Instead, it uses the mapping only to generate pseudo-measurements of the latent MFEs, which are assimilated using Kalman filtering in a controlled and modular fashion. This ensures numerical stability and consistent uncertainty quantification throughout the diagnosis-to-prognosis pipeline.

A notable exception to forward solving the corrected EoM in literature is the Bayesian-filtering-based prognosis method proposed by Subramanian et al. \cite{subramanian2023probabilistic}, which introduces a learned nonlinear mapping from system states to MFEs as a measurement model within the filtering framework. In their approach, MFEs are modeled as random walks, and an ensemble of ANNs is trained to construct a nonlinear pseudo-measurement equation. This formulation requires the use of nonlinear filters even when the nominal system dynamics is linear (because of the nonlinear measurement equation). Furthermore, the tuning of the covariance of the random walk model is not easy before knowing the response variation. The present work addresses these limitations by integrating a state-to-MFE Bayesian neural network within a GPLFM framework, enabling prognosis via standard Kalman filtering with BNN-generated pseudo-measurements, offering a clearer, modular, probabilistic alternative. 

\subsection{Overview of proposed approach}
This study develops an \textit{end-to-end probabilistic framework} for state prediction in dynamical systems governed by misspecified physics. The proposed method integrates GPLFM with a BNN to enable uncertainty-aware diagnosis, mapping, and prognosis.

In the \textit{diagnosis} phase, GPLFM is employed to jointly estimate the full system states and latent MFEs. Building on the ability to embed stationary GPs into a state-space formulation, GPLFM allows for inference via Kalman filtering and smoothing. MFEs are modeled as latent restoring forces and are estimated probabilistically alongside system states using acceleration measurements and the misspecified governing equations.

In the \textit{mapping} phase, a BNN is trained to learn the relationship between inferred states and MFEs in a probabilistic setting. The training data comprises of samples from the joint posterior distribution of states and MFEs obtained from the diagnosis step. The BNN captures epistemic uncertainty arising from the diagnosis phase and facilitates uncertainty propagation into the prognosis.

The \textit{prognosis} phase introduces a novel reuse of the GPLFM framework to perform state prediction under new excitations, but without access to actual sensor measurements. The core idea is to leverage the BNN mapping to generate pseudo-measurements: for a given input excitation, the augmented state-space system is forward-solved using the Kalman filter, where the MFEs are predicted using the BNN and treated as pseudo-measurements. This enables closed-loop Bayesian filtering even in the absence of real observations, allowing uncertainty propagation through the predictive pipeline.

\begin{figure}[H]
	\centering
	\includegraphics[scale=0.68]{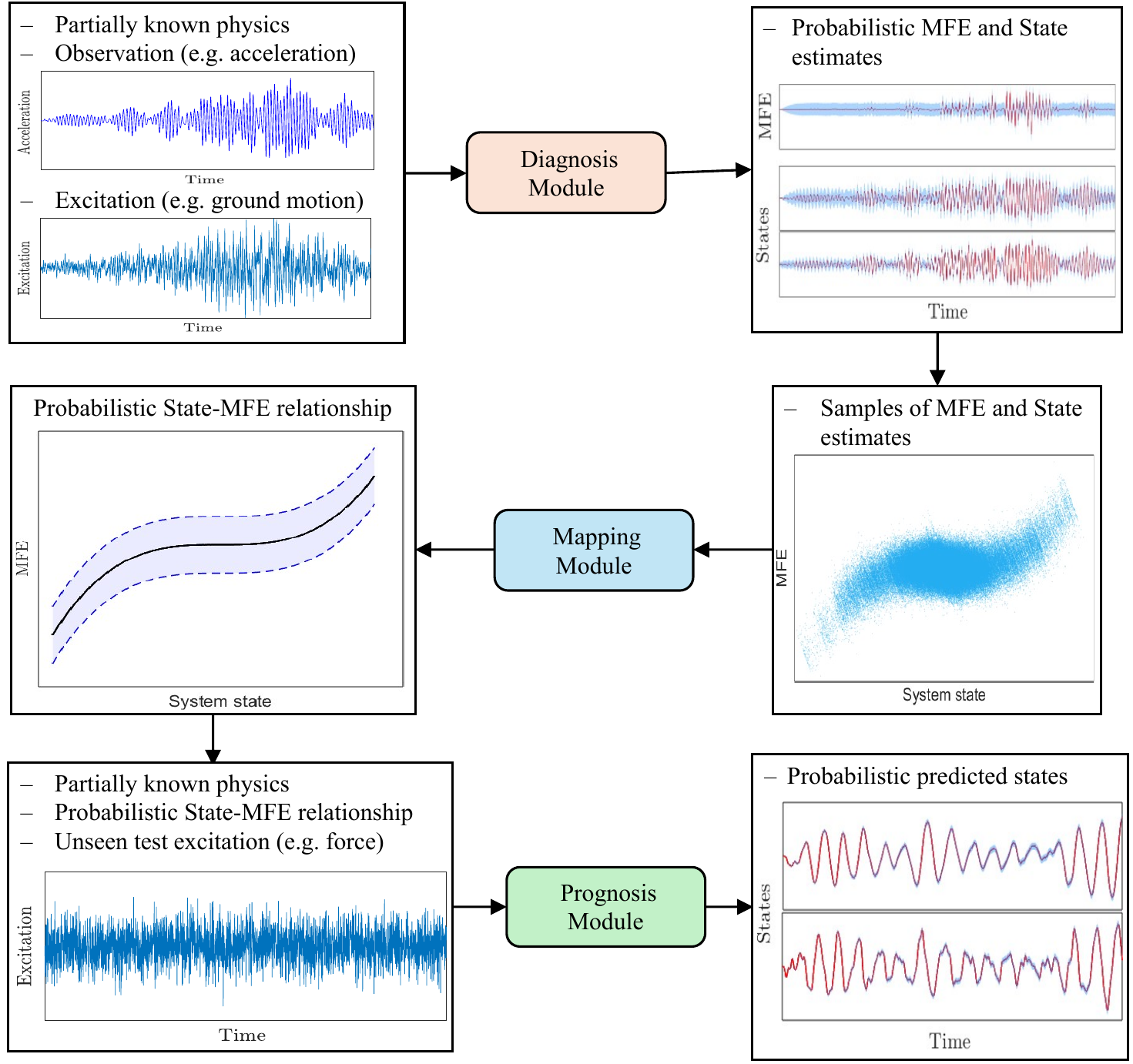}
	\caption{Schematic of the proposed framework showing the flow of data and information between stages.} 
        \label{fig: Semantic}
\end{figure}

A schematic of the overall framework is shown in \cref{fig: Semantic}. The key innovations of the proposed approach are:
\begin{itemize}
    \item An end-to-end probabilistic framework that spans diagnosis to prognosis under misspecified physics;
    \item Treatment of MFEs as latent input forces inferred via GPLFM;
    \item Use of BNN for mapping system states to MFE samples in a probabilistic setting;
    \item Use of the Kalman filter for prediction in an MFE-corrected linear system;
    \item A novel prognosis mechanism using GPLFM and BNN-generated pseudo-measurements;
    \item Explicit propagation of diagnostic uncertainty through probabilistic mapping to final predictions.
\end{itemize}

The remainder of this paper is organized as follows. \cref{sec: Prob descrip} presents the problem formulation and outlines the structure of the misspecified dynamical system. \cref{sec: Method} details the methodology: the diagnosis framework based on GPLFM is discussed in \cref{subsec: Est}, followed by the BNN-based mapping of system states to MFEs in \cref{subsec: mapping}. \cref{subsec: Pred} describes the prognosis framework using the augmented state-space model and the procedure for pseudo-measurement generation. \cref{sec: Numerical studies} demonstrates the proposed approach through numerical case studies, including two benchmark studies. \cref{sec:Discussion} offers a detailed discussion on key advantages and limitations of the proposed work. Finally, \cref{sec:conclusion} concludes with a summary and directions for future research.

\section{Problem Formulation} \label{sec: Prob descrip} 
We consider a nonlinear shear-storey-type multi-degree-of-freedom (MDOF) dynamical system with $n$ DOFs as the true physical system. Its governing equation of motion (EoM) under ground excitation $\ddot{u}_g (t)$ and external forcing $\ve{u}(t)$ is given by:
\begin{align}\label{eq:EoM_true_system_local_non-linearity}
\matr{M} \ddot{\bar{\ve{q}}} (t) + \matr{C} \dot{\bar{\ve{q}}} (t) + \matr{K} \bar{\ve{q}} (t) + \mSp \ve{p}\brc{\bar{\ve{q}} (t), \dot{\bar{\ve{q}}} (t)} = \mSu \ve{u}(t) -\matr{M} \ve{1} \ddot{u}_g(t) 
\end{align}
where $\bar{\ve{q}} (t) \in \mathbb{R}^n$ denotes the displacement vector (relative to the ground displacement ${u}_g (t)$), and $\matr{M}$, $\matr{C}$, $\matr{K}$ $\in \mathbb{R}^{n \times n}$ are the mass, damping, and stiffness matrices, respectively. The nonlinear restoring forces, which depend on displacement and velocity, are represented by $\ve{p}(\cdot)$, and applied through the matrix $\mSp$. The external force vector $\ve{u}(t)$ acts through the input influence matrix $\mSu$. 

In practice, the exact form of nonlinearities is unknown or too complex to model directly. Therefore, a simplified, nominal model is typically used, which ignores the nonlinear internal restoring force term. This results in a linear approximation of the true system by a simplified nominal model:
\begin{align} \label{eq: nominal linear model}
 \matr{M} \ddot{\breve{\ve{q}}} (t) + \matr{C} \dot{\breve{\ve{q}}} (t) + \matr{K} \breve{\ve{q}} (t) = \mSu \ve{u}(t) -\matr{M} \ve{1} \ddot{u}_g(t) 
\end{align}
This model neglects the nonlinear term  $\mSp \ve{p}(\cdot)$, resulting in model-form error (MFE) and a mismatch between the nominal response $\breve{\ve{q}} (t)$ and true response $\bar{\ve{q}} (t)$. To compensate for this discrepancy, we introduce a latent force (LF) $\ve{\eta}(t)$ as an additive correction term to the nominal model:
\begin{align} \label{eq:corrected_mdof_EoM}
\matr{M} \ddot{\ve{q}} (t) + \matr{C} \dot{\ve{q}} (t) + \matr{K} \ve{q}(t)  = \mSu \ve{u}(t)  - \matr{M} \ones \ddot{u}_g(t) - \mSp \ve{\eta}(t)
\end{align}
Subtracting this corrected model \cref{eq:corrected_mdof_EoM} from the true system \cref{eq:EoM_true_system_local_non-linearity} and assuming the corrected model tracks the true states closely (i.e., $\ve{q} (t) \approx \bar{\ve{q}} (t)$), the latent force $\ve{\eta}(t)$ approximately represents the unknown nonlinear internal restoring force:
\begin{align}
    \ve{\eta}(t) \approx \ve{p}\brc{{\ve{q}} (t), \dot{{\ve{q}}} (t)}
\end{align}
We define the state vector as the concatenation of displacements and velocities: $\ve{x}(t) = \sbk{\ve{q}^T(t)\; \; \dot{\ve{q}}^T(t)}^T$. Rewriting the second-order system into first-order form, the state-space equations consist of a continuous-time process equation and a discrete-time measurement model.
\begin{align} \label{eq:state_space_form}
    \begin{split}
    \dot{\x}(t) &= {\matr{A}_c} \x (t)
    		+
    		{\matr{B}_{pc}} \ve{\eta} (t)
                +
                {\matr{B}_{uc}} \ve{u} (t)
    		+
    		{\matr{B}_{gc}} \ddot{u}_g (t) 
    		 \\
     \ve{y}_k &= \matr{H} \ve{x}(t_k) + \matr{J}_p \ve{\eta}(t_k) + \matr{J}_u \ve{u}(t_k) + \ve{v}_k    
    \end{split}	
\end{align}
The process equation includes contributions from external input force, ground motion, and the internal latent restoring force. The measurement equation relates the sensor measurements, $\ve{y}_k \coloneq \ve{y}(t_k)$, sampled at discrete time instants $k=1, \ldots, N_t$, to the system states, LFs, and known inputs, with additive zero-mean Gaussian noise $\ve{v}_k$ with known covariance matrix $\matr{R}$. 

The system matrices $\matr{A}_c$, $\matr{B}_{uc}$, $\matr{B}_{gc}$, and $\matr{B}_{pc}$ in the state-space model are constructed from the physical parameters $\ve{\phi} = \cbk{\matr{M}, \matr{K}, \matr{C}, \matr{S}_p, \matr{S}_u}$:
\begin{align} \label{eq:SSM_process_eq_esti} 
        \matr{A}_c = \begin{bmatrix}
                \zeros & \eyei \\
                -\matr{M}^{-1} \matr{K} & -\matr{M}^{-1} \matr{C}
        \end{bmatrix}, \quad
        \matr{B}_{uc} = \begin{bmatrix}
                \zeros \\ \matr{M}^{-1} \matr{S}_u
        \end{bmatrix}, \quad
        \matr{B}_{gc} = \begin{bmatrix}
                \phantom{-}\zeros \\ -\ones 
        \end{bmatrix}, \quad
            \matr{B}_{pc} = \begin{bmatrix}
                \zeros \\ -\matr{M}^{-1} \matr{S}_p
        \end{bmatrix}  
\end{align}
Specifically, the matrix $\matr{A}_c$ captures the nominal dynamics, while $\matr{B}_{uc}$, $\matr{B}_{gc}$, and $\matr{B}_{pc}$ correspond to the influence of external forcing, ground motion, and LFs, respectively. The output matrices $\matr{H}$, $\matr{J}_p$, and $\matr{J}_u$ define how the states, LFs, and inputs contribute to the observed measurements.

Given this setup, the objective is two fold: (i) to infer the state trajectory $\ve{x}(t)$ and the LF $\ve{\eta}(t)$ from available input-output data $\mathcal{D}_{\text{obs}} = \cbk{\ve{y}_{1:N_t}, \ve{u}_{1:N_t}}$ and known model parameters $\ve{\phi}$, and (ii) to use the inferred relationship between states and LFs to predict future states $\ve{x}^\ast(t)$ under new, unseen external forces $\ve{u}^\ast(t)$, in the absence of future measurements. To this end, we define a three-stage probabilistic framework:

\begin{itemize}
    \item \textbf{Diagnosis}: Infer the posterior joint probability distribution over $\ve{x}(t)$ and $\ve{\eta}(t)$ from the observed data $\mathcal{D}_{\text{obs}}$ and known parameters $\ve{\phi}$.

    \item \textbf{Mapping}: Learn a probabilistic functional relationship $\ve{\eta}(t) \sim p \brc{\ve{\eta}(t) \mid \ve{x}(t)}$ from the inferred pairs of states and latent forces.

    \item \textbf{Prognosis}: Compute the predictive probability distribution of system states $\ve{x}^\ast(t)$ for new (previously unseen) external inputs $\ve{u}^*(t)$ using the learned mapping within the corrected nominal model.
\end{itemize}

\section{Methodology} \label{sec: Method}
To realize the proposed framework, we structure the methodology into three sequential stages: Diagnosis, Mapping, and Prognosis. The diagnosis stage estimates the MFEs (treated as LFs) and system states from observed sensor data using the probabilistic GPLFM formulation. In the mapping stage, a BNN is trained to learn a static functional relationship between the inferred states and the MFEs. The prognosis stage uses this learned probabilistic mapping within the GPLFM framework to predict the probability distribution of system states under new external force excitations.

\subsection{Diagnosis stage: Joint estimation of structural states and MFEs} \label{subsec: Est}
This section summarizes the GPLFM-based diagnosis framework used to jointly infer system states and model-form errors (MFEs). The approach builds upon our earlier work \cite{kashyap2024gaussian} and is presented here for completeness.

\subsubsection{Modeling latent forces as Gaussian processes}
The LFs $\ve{\eta}(t)$ are modeled as independent zero-mean stationary GPs, with each component $\eta_j(t)$ characterized by a covariance function (CF):
\begin{align}\label{eq: eta~GP}
    \eta_j(t) \sim  \mathcal{GP} \brc{0, \kappa^{(j)}(t-t')}, \quad j = 1, \ldots, n
\end{align}
where $\kappa^{(j)}$ is an exponential kernel defined as:
\begin{align}
    \kappa^{(j)}(t - t') = \alpha_j \exp \brc{-\frac{|t - t'|}{\ell_j}}
\end{align}
Here, $\ell_j$ denotes the lengthscale and $\alpha_j$ the signal variance. The exponential kernel admits a linear time-invariant (LTI) stochastic differential equation (SDE) representation \cite{hartikainen2012sequential, solin2019}. enabling joint inference within a state-space framework.

\subsubsection{State-space representation and augmented system}
We begin with the corrected nominal dynamics from \cref{eq:corrected_mdof_EoM}, reformulated into a continuous-time first-order state-space model. The structural dynamics and GP dynamics are represented as:
\begin{align}
\begin{split}\label{eq: corrected EoM SSM}
     \dot{\ve{x}}(t) &= \matr{A}_c \ve{x}(t) + \matr{B}_{pc} \ve{\eta}(t) + \matr{B}_{uc} \ve{u}(t) + \matr{B}_{gc} \ddot{u}_g(t) +  \ve{w}^s(t)
\end{split}
\end{align}
\begin{align}\label{eq:eta-GPSDE}
\begin{split}
        \dot{\eta}_1(t) &= - \frac{1}{\ell_j} \eta_j(t) + W_1(t), \\
        & \vdots \\
        \dot{\eta}_n(t) &= - \frac{1}{\ell_n} \eta_n(t) + W_n(t)
\end{split}
\end{align}
The Gaussian white noise terms $\ve{w}^s(t)$ and $W_j(t), \; j=1, \ldots, n$ represent process noise, with the covariance of $\ve{w}^s(t)$ set to a small value (e.g., $\sim 10^{-14}$), to prioritize the GP in capturing the modeling errors. Each $W_j(t)$ has a spectral density of $\sigma_{w,j} = \frac{2 \alpha_j}{\ell_j}$, consistent with the Mat\'{e}rn-1/2 class of GP kernels. Although we use Mat\'{e}rn-1/2 class of GP priors, any choice of stationary GP kernel can be used.

Combining the dynamics into an augmented state vector $\ve{z}(t) = \sbk{\ve{x}^T(t)\; \; \ve{\eta}^T(t)}^T$ along with the measurement equation from \cref{eq:state_space_form}, we obtain: 
\begin{align}\label{eq:augSSM_matrix_form}
\ve{\dot{z}}(t) &= \matr{F}_c\ve{z}(t) + \matr{B}^a_{uc}\ve{u}(t) + \matr{B}^a_{gc}\ddot{u}_g(t) + \ve{w}(t)
\end{align}
\begin{align}\label{eq:MeasaugSSM_matrix_form}
\ve{y}_k &= \matr{H}^a \ve{z}(t_k) + \matr{J}_u \ve{u}(t_k) + \ve{v}_k
\end{align}
where the augmented parameter matrices are defined as:
\begin{align}\label{eq: diag_aug_matrix_def}
    \matr{F}_c = 
	\begin{bmatrix}
		\matr{A}_c & \matr{B}_{pc} \\ \zeros & -\matr{\Lambda}
	\end{bmatrix}, \;
 \matr{B}^a_{uc} = 
        \begin{bmatrix}
		\matr{B}_{uc} \\ \zeros
	\end{bmatrix},\;
\matr{B}^a_{gc} =  
        \begin{bmatrix}
		\matr{B}_{gc} \\ \zeros
	\end{bmatrix}, \;
 \ve{w} = 
        \begin{bmatrix}
		{\ve{w}}^s \\ \ve{W}
	\end{bmatrix}, \;
 \matr{H}^a = \begin{bmatrix}
		\matr{H} & \matr{J}_p
	\end{bmatrix}
\end{align}
Here $\matr{\Lambda} = \text{diag} \cbk{\frac{1}{\ell_1}, \ldots, \frac{1}{\ell_n}}$ is a $n\times n$ positive-definite matrix and $\ve{W} = \cbk{W_1, \dots, W_n}$ is a $n \times 1$ column vector of independent white noise processes with corresponding spectral densities, $\sigma_{w,1}, \ldots, \sigma_{w,n}$, and $\ve{w}$ is a zero-mean Gaussian white noise vector with covariance $\matr{Q}_c = \texttt{blkdiag}(10^{-14} \mathbb{I}_n,\sigma_{w,1}, \ldots, \sigma_{w,n})$; $\mathbb{I}_n$ is a square identity matrix of size $n \times n$. Definitions of all matrices such as $\matr{A}_c$, $\matr{B}_{uc}$, etc.\ follow the definitions provided in Section~\ref{sec: Prob descrip}.

\subsubsection{MAP estimation of GP hyperparamters} \label{subsec: optim Est.}
The GP hyperparameters $\vtheta = \cbk{\ell_j, \alpha_j}_{j=1}^n$ are estimated via maximum \textit{a posteriori} (MAP) inference:
\begin{align}\label{eq: opti_diagnosis}
    \hat{\vtheta}_{\text{MAP}} = \underset{\vtheta}{\arg\min} \left( - \log \pr{\ve{y}_{1:N_t} \mid \vtheta, \ve{\phi}, \ve{u}_{1:N_t}} - \log \pr{\vtheta} \right)
\end{align}
Given the physical parameters $\ve{\phi}$ and external force inputs $\ve{u}_{1:N_t}$, the marginal likelihood $p \brc{\ve{y}_{1:N_t} \mid \vtheta, \ve{\phi}, \ve{u}_{1:N_t}}$ is evaluated using the Kalman filter recursion. \textcolor{black}{When the specific DOFs affected by MFEs are unknown \textit{apriori}, it becomes crucial to automatically localize the DOFs where MFEs are present. In practice, if LFs are introduced at all DOFs, the algorithm should ideally drive the LFs at MFE-free DOFs toward zero while retaining non-zero LFs only at DOFs associated with MFEs. This behavior can be encouraged by enforcing sparsity in the LF estimates through appropriate priors on the GP hyperparameters.}

\textcolor{black}{To promote such sparsity, we impose a Student’s-$t$ prior distribution on the GP hyperparameters. A DOF with no MFE would exhibit an approximately constant (zero) LF time series, corresponding to $\alpha \!\approx\! 0$ (negligible signal magnitude) and large $\ell$ (slow variation). Conversely, a DOF with an MFE would yield an LF with large $\alpha$ and small $\ell$. The heavy tails of the Student’s-$t$ prior provide the necessary flexibility to accommodate both cases, allowing hyperparameters to take values far from their mean when required by the data. In this work, the statistical degree-of-freedom parameter of the Student ’s-$t$ prior is fixed to $\nu = 1$ (corresponding to a Cauchy prior). This choice introduces the heaviest tails, enabling large excursions in the hyperparameters if needed to represent MFEs, while strongly shrinking irrelevant LFs toward zero. The value of $\nu$ is not optimized, as it is specifically chosen to enforce heavy-tailed regularization rather than to fit the data. Further details on the GP hyperparameter optimization are provided in \cref{appendix: GP hyperparameter Opti}.}

\subsubsection{Joint inference of states and MFEs}
Given the MAP estimate $\hat{\vtheta}$, the augmented state posterior $p\brc{\ve{z}_{1:N_t} \mid \hat{\vtheta}, \ve{\phi}, \mathcal{D}_{\text{obs}}}$ is inferred using the Kalman filter (KF) \cite{kalman1960new} and the Rauch-Tung-Striebel (RTS) smoother \cite{rauch1965maximum}. This yields smoothed estimates and covariances for both system states and LFs, which are subsequently used in the mapping phase, Section \ref{subsec: mapping}. Details of the KF and RTS steps are provided in \ref{appendix:Kfs}. The diagnosis stage is summarized in \cref{algo: Diagnosis}.

\begin{algorithm}
\caption{\textbf{Diagnosis}: Joint inference of structural states and MFEs using GPLFM}
\label{algo: Diagnosis}
\begin{algorithmic}[1]
\Statex \textbf{Input:} Observed input-output data $\mathcal{D}_{\text{obs}} = \{\ve{y}_{1:N_t}, \ve{u}_{1:N_t}\}$, initial condition $\ve{x}_0$, and structural parameters $\ve{\phi} = \{\matr{M}, \matr{K}, \matr{C}, \matr{S}_p, \matr{S}_u\}$
\Statex \textbf{Output:} Joint posterior distribution $\pr{\ve{z}_{1:N_t} \mid \hat{\vtheta}, \ve{\phi}, \mathcal{D}_{\text{obs}}}$ over augmented state $\ve{z}_k = \begin{bmatrix} \ve{x}_k^T & \ve{\eta}_k^T \end{bmatrix}^T$

\State Construct the continuous-time state-space model for the latent-force-corrected EoM as per \cref{eq:state_space_form}, incorporating model-form errors as additive forces.

\State Model the MFEs $\ve{\eta}(t)$ as a stationary GP, and represent it using an SDE formulation as in \cref{eq:eta-GPSDE}, with an initial guess of GP hyperparameters $\vtheta$ (i.e., length scale, signal variance).

\State Form the continuous-time augmented state vector $\ve{z}_k = \begin{bmatrix} \ve{x}_k^T & \ve{\eta}_k^T \end{bmatrix}^T$ and define the augmented process and measurement models as in \cref{eq:augSSM_matrix_form} and \cref{eq:MeasaugSSM_matrix_form}, with augmented system matrices $\matr{F}_c$, $\matr{B}_{uc}^a$, $\matr{B}_{gc}^a$, process noise ${\ve{w}}$, and observation matrix $\matr{H}^a$ specified in \cref{eq: diag_aug_matrix_def}.

\State Discretize the continuous-time augmented system using a suitable numerical scheme (e.g., zero-order hold) to obtain the discrete-time process model \cref{eq:disc_SSM}.

\State Estimate the optimal GP hyperparameters $\hat{\vtheta}$ by maximizing the marginal posterior using measurement data $\ve{y}_{1:N_t}$, as defined in the MAP objective in \cref{eq: opti_diagnosis}.

\State Apply Kalman filter followed by RTS smoother to obtain the smoothed joint posterior distribution over states and MFEs:
$\pr{\ve{x}_{1:N_t}, \ve{\eta}_{1:N_t} \mid \hat{\vtheta}, \ve{\phi}, \ve{u}_{1:N_t}, \ve{y}_{1:N_t}}$

\end{algorithmic}
\end{algorithm}

Having obtained a joint posterior distribution over system states and MFEs from the diagnosis stage, we now aim to construct a functional relationship between them. Specifically, we hypothesize that the MFEs --- modeled as latent forces during diagnosis --- exhibit a direct, potentially nonlinear dependence on the system state. Learning this relationship enables us to generalize the latent behavior observed during training and subsequently use it for predicting future evolution under unseen excitations. The next stage, therefore, involves learning a probabilistic mapping from the system state $\ve{x}$ to the MFE $\ve{\eta}$ using a Bayesian Neural Network (BNN), which captures both the mean trend and the state-dependent uncertainty inherent in this relationship.

\subsection{Mapping stage: Learning a probabilistic static map}\label{subsec: mapping}
Following the diagnosis phase, where the system states and MFEs are jointly estimated, we now construct a static probabilistic mapping between the two. Specifically, the objective is to learn the conditional distribution $p\brc{\ve{\eta} \mid \ve{x}}$ using the posterior samples obtained from the diagnosis phase. This mapping will later be used to generate pseudo-measurements for probabilistic prognosis. We assume an instantaneous and explicit dependence, i.e., $\ve{\eta}(t_k)$ depends only on the current state $\ve{x}(t_k)$ and not its history. While this assumption may not strictly hold in all scenarios --- particularly for systems with history-dependent nonlinearities --- it still offers an effective approximation (as is demonstrated in one of the examples in Section \ref{sec: Numerical studies}). 

\subsubsection{Training data from posterior samples}
The diagnosis stage yields a smoothed joint posterior  $p\brc{\ve{x}_{1:N_t}, \ve{\eta}_{1:N_t} \mid \ve{y}_{1:N_t}, \ve{u}_{1:N_t}, \ve{\phi}}$, from which we draw samples of paired realizations $\brc{\ve{x}^{(i)}, \ve{\eta}^{(i)}}$ at each time step. Aggregating these samples across time yields a training dataset $\mathcal{D}_{\text{map}} = \cbk{ \brc{\ve{x}^{(i)}, \ve{\eta}^{(i)}} }_{i=1}^{N_m}$, with $\ve{x}$ as input and $\ve{\eta}$ as output. Typically, $N_m \gg N_t$, as multiple samples can be drawn per time step, enabling a richer learning of the conditional distribution $p(\ve{\eta} \mid \ve{x})$.

Importantly, both inputs and outputs are sampled from a posterior distribution, propagating epistemic uncertainty from the diagnosis phase into the learning process. Since the mapping is learned from time-indexed samples but assumed to be static (i.e.\ time-invariant), the training distribution becomes heteroskedastic --- the conditional variance of $\ve{\eta}$ varies across $\ve{x}$ due to the time-varying posterior uncertainty in the joint posterior distribution from diagnosis phase.

\begin{figure}[H]
	\centering
\includegraphics[width=\textwidth]{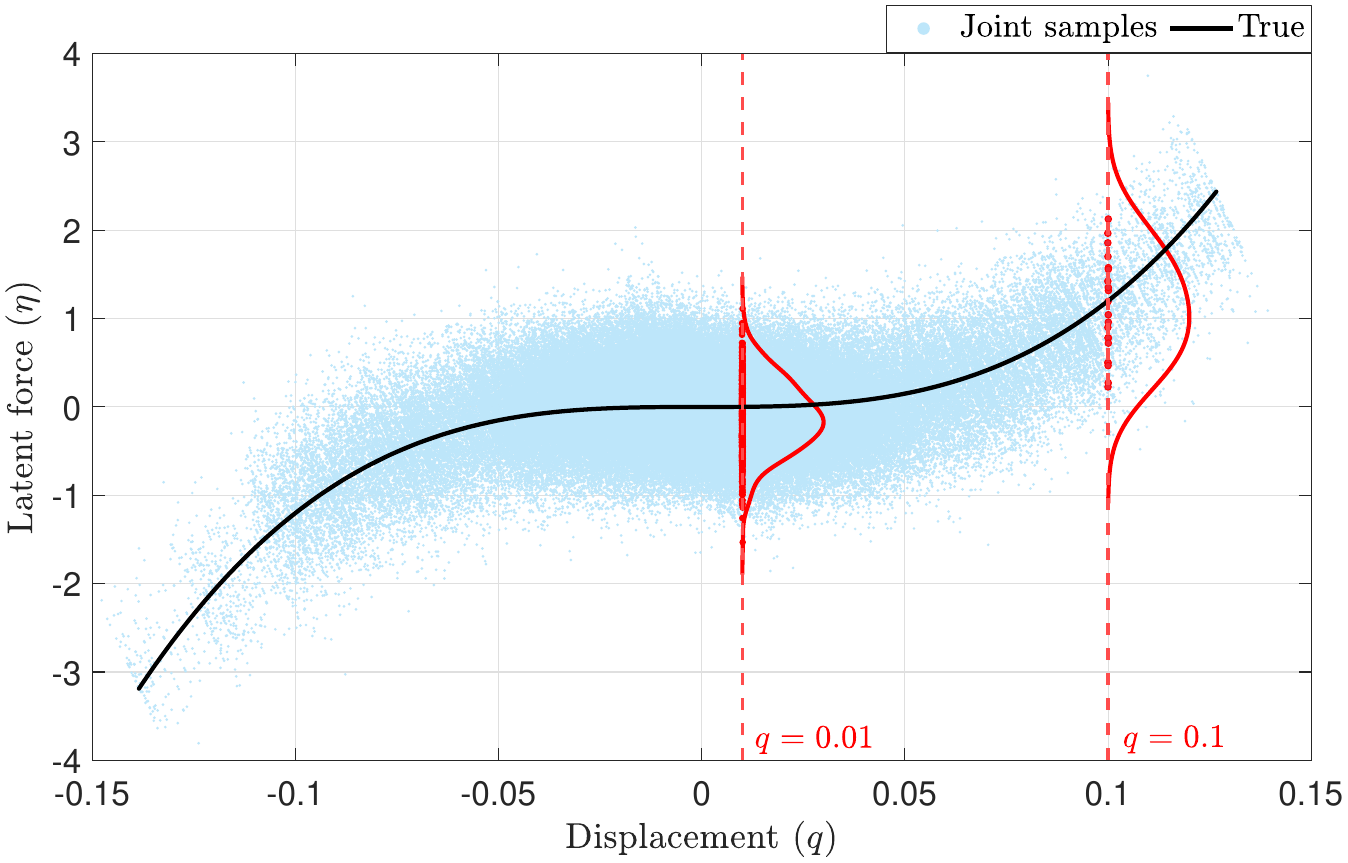}
    \caption{Posterior samples of displacement $q$ versus MFE $\eta$, drawn from $p(\ve{x}_k, \eta_k \mid \ve{y}_{1:N_t})$ at each time step. In this case, $\eta$ represents an unmodelled cubic displacement nonlinearity $K_{\text{nl}}q^3$, where $K_{\text{nl}}$ is a constant. Vertical dashed lines indicate selected values of $q$ (e.g., $q = 0.01$, $0.1$), along which the corresponding conditional distributions $\pr{\eta \mid q}$ are shown. The associated $\eta$ samples are highlighted in matching colors. The nonlinear trend and variation in spread of $\eta$ across $q$ illustrate the heteroskedastic and state-dependent nature of $p(\eta \mid \ve{x})$, motivating the use of BNNs for mapping.}
        \label{fig: ensemble_x_vs_eta_Duffing}
\end{figure}

This behavior is illustrated in Figure~\ref{fig: ensemble_x_vs_eta_Duffing}, where an ensemble of samples drawn from a joint posterior for an SDOF system with cubic displacement nonlinearity --- treated as an MFE --- shows both a clear nonlinear trend and state-dependent spread in the MFE $\eta$. Such characteristics motivate the use of a BNN, which naturally models both the nonlinear trend and the input-dependent uncertainty.

\subsubsection{BNN architecture and output parameterization}
We model the conditional distribution $p(\ve{\eta} \mid \ve{x})$ using a BNN. Given input $\ve{x} $, the BNN predicts both the mean $\ve{\mu}^\eta(\ve{x}, \ve{\varphi})$ and the covariance $\matr{\Sigma}^\eta(\ve{x}, \ve{\varphi})$ of a Gaussian likelihood as:
\begin{align}
    p(\ve{\eta} \mid \ve{x}, \ve{\varphi}) = \mathcal{N}\brc{\ve{\eta} \mid \ve{\mu}^{\eta}(\ve{x}, \ve{\varphi}), \matr{\Sigma}^{\eta}(\ve{x}, \ve{\varphi})}
\end{align}
Here, $\ve{\varphi}$ represents the set of neural network parameters. To ensure $\matr{\Sigma}^{\eta}$ is symmetric and positive definite, the network predicts the elements of a lower triangular matrix $\matr{L}(\ve{x},\ve{\varphi})$, such that $\matr{\Sigma}^{\eta} = \matr{L} \matr{L}^T$. The diagonal elements of $\matr{L}$ are constrained using a softplus transformation to ensure positive variances.

Thus, for an $n$-dimensional output $\ve{\eta}$, the BNN has a total of $n + \frac{n(n+1)}{2}$ output nodes (or neurons):
\begin{itemize}
    \item $n$ for the mean vector $\ve{\mu}^\eta$,

    \item $\frac{n(n+1)}{2}$ for the Cholesky factor $\matr{L}$.
\end{itemize}

\begin{figure} [H]
\centering
\includegraphics[scale=0.55]{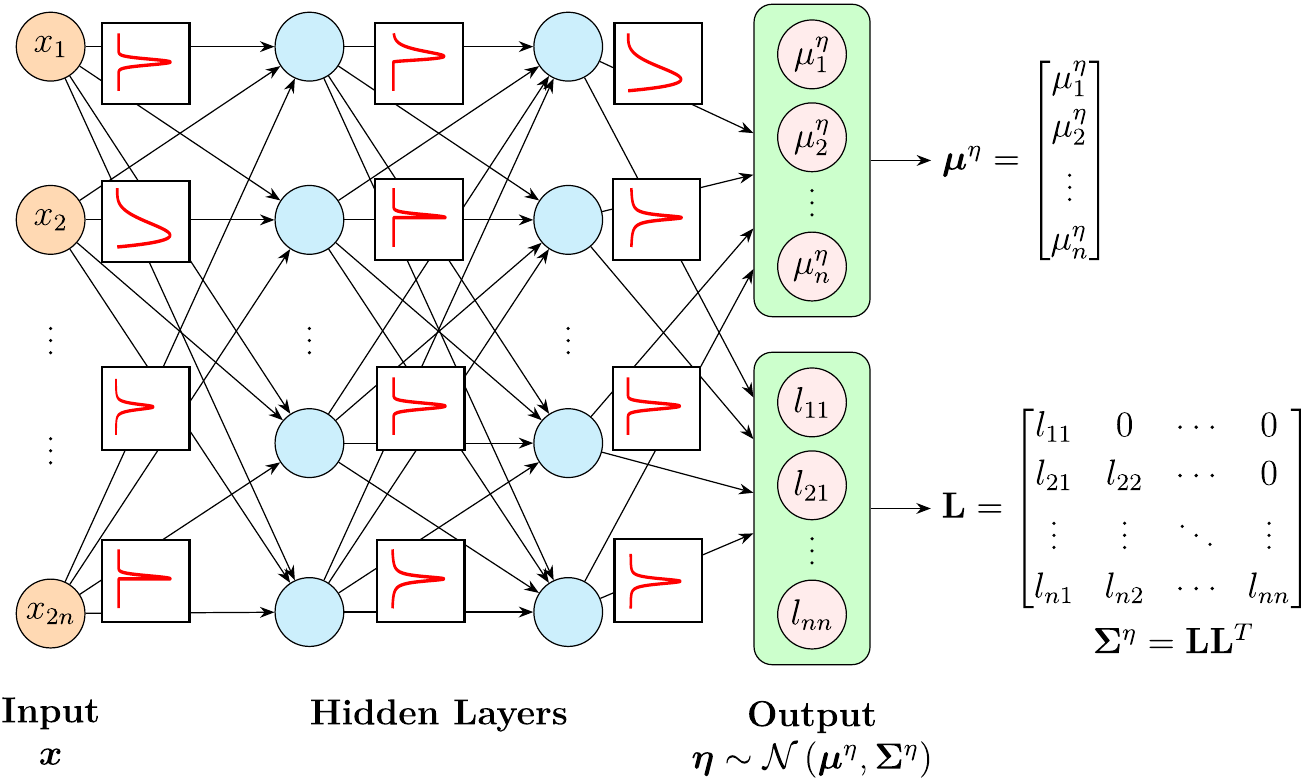}
\caption{Schematic of the BNN architecture mapping system states $\ve{x}$ to model-form errors $\ve{\eta}$. The output layer predicts the mean $\ve{\mu}^\eta$ and the Cholesky factor $\matr{L}$ to reconstruct the full covariance $\matr{\Sigma}^\eta = \matr{L} \matr{L}^T$.}
\label{fig: BNN_sch}
\end{figure}

Further, a standard Gaussian prior is placed over the network parameters:
\begin{align}
p(\ve{\varphi}) = \prod_j \mathcal{N}(\varphi_j \mid 0, 1)
\end{align}
This choice regularizes the network, encouraging weights to remain small unless supported by the data.
\subsubsection{Training the BNN via variational inference}
The true posterior distribution over weights $p(\ve{\varphi} \mid \mathcal{D})$ is intractable and is approximated using mean field variational inference. Specifically, a variational distribution in the form of a fully factorized Gaussian distribution is assumed over the network parameters:
\begin{align}
    q(\ve{\varphi}) = \prod_j \mathcal{N}\brc{\varphi_j \mid \mu^q_j, \brc{\sigma_j^q}^2}
\end{align}
The variational parameters $\mu_j^q$ and $\sigma_j^q$ are optimized by minimizing the negative \textit{evidence lower bound} (ELBO):
\begin{align} \label{eq: Loss_BNN}
    \mathcal{L}_{\text{VI}} = -\sum_{i=1}^{N_m} \mathbb{E}_{q(\ve{\varphi})} \sbk{ \log p \brc{\ve{\eta}^{(i)} \mid \ve{x}^{(i)}, \ve{\varphi}} } + \mathrm{KL}\sbk{q(\ve{\varphi}) || p(\ve{\varphi})}
\end{align}
Here, the first term encourages the BNN to match observed data via the expected log-likelihood, while the second term regularizes the network by penalizing deviation from the prior.

The expectation over the log-likelihood is approximated using Monte Carlo sampling by drawing multiple realizations $\ve{\varphi}^{(s)} \sim q(\ve{\varphi})$. The KL divergence term admits a closed-form expression when both $q$ and $p$ are Gaussians (see \ref{appendix: KLD}).

In practice, optimization is performed using stochastic mini-batches to handle large datasets efficiently. Let $\mathcal{B}^{(k)} \subset \mathcal{D}_{\text{map}}$ denote the $k^\text{th}$ mini-batch of size $B$ in the current epoch. Then, the ELBO is approximated as:
\begin{align}\label{eq: BNN_loss_batch}
\mathcal{L}_{\text{VI}}^{(k)} \approx -\frac{1}{B} \sum_{b=1}^{B} \sbk{ \frac{1}{S} \sum_{s=1}^{S} \log \pr{ \ve{\eta}^{(b)} \mid \ve{x}^{(b)}, \ve{\varphi}^{(s)} } } + \mathrm{KL}\sbk{q(\ve{\varphi}) || p(\ve{\varphi})}
\end{align}
where $S$ is the number of Monte Carlo samples used to approximate the expected log-likelihood.

Training proceeds iteratively by updating the variational parameters based on the loss computed in each mini-batch. One iteration refers to the processing of a single mini-batch, while an epoch refers to a full pass through the entire dataset. Convergence is monitored by tracking the ELBO at the end of each ${e}^\text{th}$ epoch, and training is terminated when the change in ELBO falls below a predefined threshold:
\begin{align} \label{eq: BNN_converge}
\left| \mathcal{L}_{\text{VI}}^{(e)} - \mathcal{L}_{\text{VI}}^{(e-1)} \right| < \varepsilon
\end{align}
where $\varepsilon$ is a small positive scalar (e.g., $10^{-4}$) controlling convergence sensitivity.

\subsubsection{Posterior predictive distribution}
After training the BNN, the posterior predictive distribution for a new input $\ve{x}^\ast$ is obtained by marginalizing over the optimized variational posterior $q^\ast(\ve{\varphi})$:
\begin{align}
\pr{\ve{\eta} \mid \ve{x}^\ast, \mathcal{D}_{\text{map}}} \approx \int p(\ve{\eta} \mid \ve{x}^\ast, \ve{\varphi}) \;\; q^\ast \brc{\ve{\varphi} \mid \mathcal{D}_{\text{map}}}  d\ve{\varphi}
\label{eq:pred_bnn_integral}
\end{align}
where, the likelihood $p(\ve{\eta} \mid \ve{x}^\ast, \ve{\varphi})$ is a multivariate Gaussian with mean $\ve{\mu}^\eta(\ve{x}^\ast, \ve{\varphi})$ and covariance $\matr{\Sigma}^\eta(\ve{x}^\ast, \ve{\varphi})$ parameterized by the BNN.
To evaluate this integral, we use Monte Carlo sampling. Specifically, we draw $S$ samples $\ve{\varphi}^{(s)} \sim q^\ast(\ve{\varphi})$, and compute the corresponding predicted mean and covariance:
\begin{align*}
\ve{\mu}^{(s)} = \ve{\mu}^\eta \brc{\ve{x}^\ast, \ve{\varphi}^{(s)}}, \quad 
\matr{\Sigma}^{(s)} = \matr{\Sigma}^\eta \brc{\ve{x}^\ast, \ve{\varphi}^{(s)}}
\end{align*}
The predictive mean $\bar{\ve{\mu}}^\eta$ and covariance $\bar{\matr{\Sigma}}^\eta$ are then approximated by moment-matching:
\begin{align}
\bar{\ve{\mu}}^\eta &= \frac{1}{S} \sum_{s=1}^S \ve{\mu}^{(s)} \label{eq:mean_pred_eta} \\
\bar{\matr{\Sigma}}^\eta &= \frac{1}{S} \sum_{s=1}^S \matr{\Sigma}^{(s)} + \frac{1}{S} \sum_{s=1}^S \left( \ve{\mu}^{(s)} - \bar{\ve{\mu}}^\eta \right) \left( \ve{\mu}^{(s)} - \bar{\ve{\mu}}^\eta \right)^T \label{eq:cov_pred_eta}
\end{align}
This results in a multivariate Gaussian approximation:
\begin{align}
\pr{\ve{\eta} \mid \ve{x}^\ast, \mathcal{D}_{\text{map}} } \approx \mathcal{N}\left( \ve{\eta} \mid \bar{\ve{\mu}}^\eta, \bar{\matr{\Sigma}}^\eta \right)
\label{eq:pred_bnn_gaussian}
\end{align}
This predictive distribution is subsequently used to generate pseudo-measurements in the Kalman filtering step of prognosis.

\begin{algorithm}
\caption{\textbf{Mapping}: Learning probabilistic state-to-MFE map using BNN}
\label{algo: MappingBNN}
\begin{algorithmic}[1]
\Statex \textbf{Input:} Joint posterior $p(\ve{z}_{1:N_t} \mid \hat{\vtheta}, \ve{\phi}, \mathcal{D}_{\text{obs}})$ from diagnosis stage, where $\ve{z}_k = \sbk{\ve{x}_k^T \; \ve{\eta}_k^T}^T$
\Statex \textbf{Output:} Posterior predictive distribution $p(\ve{\eta} \mid \ve{x}, \mathcal{D}_{\text{map}})$ used in prognosis
\vspace{0.3em}

\Statex \textbf{Step 1: Training data preparation}
\State Draw $N_m$ samples $\cbk{ \brc{\ve{x}^{(i)}, \ve{\eta}^{(i)}} }_{i=1}^{N_m}$ from $p(\ve{z}_{1:N_t} \mid \hat{\vtheta}, \ve{\phi}, \mathcal{D}_{\text{obs}})$
\State Form training dataset $\mathcal{D}_{\text{map}} = \cbk{ \brc{\ve{x}^{(i)}, \ve{\eta}^{(i)}} }_{i=1}^{N_m}$

\vspace{0.3em}
\Statex \textbf{Step 2: BNN definition}
\State Define network output as Gaussian likelihood: $p(\ve{\eta} \mid \ve{x}, \ve{\varphi}) = \mathcal{N}(\ve{\mu}^\eta(\ve{x}, \ve{\varphi}), \matr{\Sigma}^\eta(\ve{x}, \ve{\varphi}))$
\State Assign standard normal prior on weights: $p(\ve{\varphi}) = \prod_j \mathcal{N}(0, 1)$
\State Define variational posterior: $q(\ve{\varphi}) = \prod_j \mathcal{N}(\mu^q_j, (\sigma^q_j)^2)$

\vspace{0.3em}
\Statex \textbf{Step 3: Variational training}
\State Initialize variational parameters $\cbk{\mu^q_j, \sigma^q_j}$
\For{each epoch $e = 1$ to $E$}
    \For{each mini-batch $\mathcal{B}^{(k)} = \cbk{(\ve{x}^{(b)}, \ve{\eta}^{(b)})}_{b=1}^B \subset \mathcal{D}_{\text{map}}$}
        \State Draw $S$ samples $\cbk{ \ve{\varphi}^{(s)} }_{s=1}^S \sim q(\ve{\varphi})$
        \State For each $\ve{\varphi}^{(s)}$, compute $\ve{\mu}^{(s)} = \ve{\mu}^\eta \brc{\ve{x}^{(b)}, \ve{\varphi}^{(s)}}$ and $\matr{\Sigma}^{(s)} = \matr{\Sigma}^\eta \brc{\ve{x}^{(b)}, \ve{\varphi}^{(s)}}$
        \State Compute minibatch ELBO loss $\mathcal{L}^{(k)}_{\text{VI}}$ as in \eqref{eq: BNN_loss_batch}
        \State Update variational parameters $\cbk{\mu^q_j, \sigma^q_j}$ using gradient-based optimizer (e.g.\ Adam)
    \EndFor
    \State Compute epoch-averaged loss $\mathcal{L}_{\text{VI}}^{(e)}$
    \State Continue iterations until convergence (checked using condition in \cref{eq: BNN_converge})
\EndFor

\vspace{0.3em}
\Statex \textbf{Step 4: Posterior predictive distribution}
\State For any new $\ve{x}^\ast$, sample $\ve{\varphi}^{(s)} \sim q^\ast(\ve{\varphi})$ and compute $\ve{\mu}^{(s)}$, $\matr{\Sigma}^{(s)}$
\State Compute predictive mean $\bar{\ve{\mu}}^\eta$ and covariance $\bar{\matr{\Sigma}}^\eta$ via moment-matching using \cref{eq:mean_pred_eta,eq:cov_pred_eta}
\State Return: $\pr{\ve{\eta} \mid \ve{x}^\ast, \mathcal{D}_{\text{map}}} \approx \mathcal{N}(\bar{\ve{\mu}}^\eta, \bar{\matr{\Sigma}}^\eta)$
\end{algorithmic}
\end{algorithm}

The learned probabilistic mapping from system states to MFEs serves as a crucial component in the final stage which is prognosis. While the diagnosis stage enables posterior inference over latent dynamics using observed data, and the mapping stage captures a generalized, uncertainty-aware relationship between $\ve{x}$ and $\ve{\eta}$, prognosis aims to predict the future evolution of the system under new, \textcolor{black}{observed but unseen (during diagnosis phases)} inputs. A key challenge here is the recursive nature of prediction, where unknown MFEs influence the state trajectory, and states in turn influence MFEs. To resolve this, we avoid embedding the learned BNN directly into the dynamics and instead use it to generate pseudo-measurements of MFEs, which are assimilated using a Kalman filter. The following section details this hybrid inference procedure, integrating Gaussian process modeling, state-space dynamics, and BNN-based measurement updates for robust and scalable state prediction.

\subsection{Prognosis stage: State prediction under new external force} \label{subsec: Pred}
The final phase of the framework focuses on the prediction of system states under new, previously unseen external force excitations $\ve{u}^\ast(t)$.  Traditional approaches often incorporate the deterministic part of the state-to-MFE mapping directly into the governing (a.k.a.\ process) equation and then apply numerical ODE solvers to obtain the predicted states \cite{garg2022physics}. These methods typically rely on the mean prediction from a learned distribution $p(\ve{\eta} \mid \ve{x})$, neglecting the uncertainty inherent in the mapping. 

A more rigorous yet computationally intensive approach to predictive modeling involves performing probabilistic forward integration of the MFE-corrected equations using Monte Carlo sampling. In this approach, realizations of both system states and MFEs are drawn at each time step and propagated through time to characterize uncertainty. However, this procedure suffers from significant drawbacks. Since the learned mapping from system states to MFEs is purely data-driven and not constrained by dynamical stability requirements, embedding it directly into the process model for numerical integration can result in unstable trajectories. In our experience, many such realizations tend to diverge due to the nonlinear and often unconstrained nature of the MFE realizations. This not only compromises the physical reliability of the predictions but also makes the approach computationally expensive and numerically fragile, limiting its practical appeal.


To overcome these challenges, we propose a more stable and tractable alternative inspired by the diagnosis phase. Specifically, we continue to treat MFEs as latent forces modeled via Gaussian processes and integrate them into an augmented linear state-space model. The key idea lies in the use of the learned BNN-based mapping $p(\ve{\eta} \mid \ve{x})$ to generate \textit{pseudo-measurements} of MFEs conditioned on predicted states, rather than embedding the nonlinear map directly into the process equation. These pseudo-measurements are assimilated using Kalman filtering, enabling recursive probabilistic inference over states while preserving the linear structure of the process equation.

Although this approach shifts some computational burden to GP hyperparameter optimization, it offers a principled and efficient pathway for uncertainty-aware state prediction. It retains the expressive power of data-driven nonlinear corrections while avoiding the numerical instabilities that often accompany direct integration of learned dynamics, thus enabling robust and stable prognosis under new excitations.

\subsubsection{Latent force modeling for new external inputs}
Since the MFEs, $\ve{\eta}(t)$, are implicitly dependent on the state trajectories --- and hence on the external force input --- the MFE must be re-specified to reflect the new external force input $\ve{u}^\ast(t)$.

To retain stability and tractability, we continue to model the latent force term $\ve{\eta}^\ast(t)$ as a GP, similar to the diagnosis phase. That is, each component $\eta^\ast_j(t)$ is modeled as an independent, stationary GP with zero mean and kernel $\kappa\brc{t - t'; \vtheta^\ast}$:
\begin{align} \label{eq:pred_eq_MKC1_GPbasic}
\eta^\ast_j(t) \sim \mathcal{GP} \brc{0, \kappa(t-t';\vtheta^\ast)}, \quad j = 1, \ldots, n,
\end{align}
where $\vtheta^\ast = \cbk{\ell^\ast_j, \alpha^\ast_j}$ are the GP kernel hyperparameters specific to the prognosis phase. The choice of a stationary GP is deliberate; it enables an equivalent LTI-SDE formulation, allowing seamless integration of MFEs into a linear augmented state-space model. This strategy ensures numerical stability during prediction and preserves the benefits of Kalman filtering.

\subsubsection{Augmented linear process equation for Kalman prediction step}
The Kalman filter carries out recursive Bayesian state estimation through a two-step process: a \textit{prediction step}, which propagates the prior over system states using the process equation, and an \textit{update step}, which incorporates new measurements (or pseudo-measurements, in our prognosis phase).

For the prediction step, we formulate an augmented state-space model that incorporates both the structural state $\ve{x}^\ast(t)$ and the latent force $\ve{\eta}^\ast(t)$. This results in the continuous-time process equation:
\begin{align} \label{eq:Process_eq_pred} 
\ve{\dot{z}}^\ast(t) = \matr{F}_c^\ast \ve{z}^\ast(t) + \matr{B}^a_{uc} \ve{u}^\ast(t) + \matr{B}^a_{gc} \ddot{u}_g(t) + \ve{w}^\ast(t),
\end{align}
with augmented state $\ve{z}^\ast(t) = \sbk{{\ve{x}^{\ast}}^T(t) \quad {\ve{\eta}^{\ast}}^T(t)}^T$. The system matrices are defined as:
\begin{align}
    \matr{F}_c^\ast = 
	\begin{bmatrix}
		\matr{A}_c & \matr{B}_{pc} \\ \zeros & -\matr{\Lambda}^\ast
	\end{bmatrix}, \;
 \matr{B}^a_{uc} = 
        \begin{bmatrix}
		\matr{B}_{uc} \\ \zeros
	\end{bmatrix},\;
\matr{B}^a_{gc} =  
        \begin{bmatrix}
		\matr{B}_{gc} \\ \zeros
	\end{bmatrix}
\end{align}
where $\matr{\Lambda}^\ast = \text{diag} \cbk{\frac{1}{\ell_1^\ast}, \ldots, \frac{1}{\ell_n^\ast}}$ is a $n\times n$ positive-definite diagonal matrix. The process noise $\ve{w}^\ast(t)$ comprises 
\begin{align}
    \ve{w}^\ast(t) = 
        \begin{bmatrix}
		{\ve{w}}^s(t) \\ \ve{W}^\ast(t)
	\end{bmatrix}, \quad \text{with   } \mathbb{E}\sbk{ \begin{bmatrix}
		{\ve{w}}^s(t) \\ \ve{W}^\ast(t)
	\end{bmatrix} \begin{bmatrix}
		{\ve{w}}^s(t') \\ \ve{W}^\ast(t')
	\end{bmatrix}^T } = \delta(t - t') \underbrace{\begin{bmatrix}
	    10^{-14} \mathbb{I}_{2n} & \zeros \\ \zeros & \matr{Q}_\eta^\ast 
	\end{bmatrix}}_{\matr{Q}_c^\ast}
\end{align}
where $\matr{Q}_\eta^\ast = \texttt{diag} \brc{\sigma_{w,1}^\ast, \ldots, \sigma_{w,n}^\ast}$ and $\sigma_{w,j}^\ast= \frac{2 \alpha_j^\ast}{\ell_j^\ast}$. Note that these noise intensities are directly linked to the GP kernel hyperparameters (to be learned via optimization).

This continuous-time model is then discretized using zero-order hold assumption, following standard discretization rules (e.g., see \cite{sarkka2013bayesian}), resulting in:
\begin{align} \label{eq:Process_eq_pred_disc}
\ve{z}^\ast_k = \matr{F}_d^\ast \ve{z}^\ast_{k-1} + \matr{B}^a_{ud} \ve{u}^\ast_{k-1} + \matr{B}^a_{gd} \ddot{u}_{g,k-1} + \ve{w}^\ast_{k-1},
\end{align}
where $\ve{w}_{k-1}^\ast \sim \mathcal{N} \brc{\zeros, \matr{Q}_d^\ast}$

The discrete-time process equation is used in the Kalman prediction step, yielding a Gaussian distribution over the augmented state at each time step:
\begin{align} \label{eq:Pred_aug_states}
\ve{z}_{k\mid k-1}^\ast \sim \mathcal{N} \brc{ \ve{m}_{k \mid k-1}^{z,\ast}, \matr{P}_{k\mid k-1}^{z,\ast} }
\end{align}
where $\ve{m}^{z,\ast}_{k\mid k-1}$ and $\matr{P}^{z,\ast}_{k\mid k-1}$ denote the predicted mean and covariance of the augmented state vector obtained using the following recursion:
\begin{align}\label{eq: Prog-KF pred}
\ve{m}^{z,\ast}_{k\mid k-1} &= \matr{F}_d^\ast \ve{m}^{z,\ast}_{k-1\mid k-1} + \matr{B}^a_{ud} \ve{u}^\ast_{k-1} + \matr{B}^a_{gd} \ddot{u}_{g,k-1} \\
\matr{P}^{z,\ast}_{k\mid k-1} &= \matr{F}_d^\ast \matr{P}^{z,\ast}_{k-1\mid k-1} \matr{F}_d^{\ast T} + \matr{Q}_d^\ast
\end{align}

\subsubsection{Measurement equation and pseudo-measurement generation}
In standard Kalman filtering, the update step relies on real measurements of system states. However, during prognosis under a new external excitation \(\ve{u}^\ast(t)\), no real measurements of the structural states are available. To enable the update step in this measurement-free regime, we introduce a novel mechanism for generating pseudo-measurements of the MFEs using the learned BNN-based mapping.

From the joint Gaussian distribution of the structural states $\ve{x}^\ast(t)$ and latent forces $\ve{\eta}^\ast(t)$ obtained in \cref{eq: Prog-KF pred}, we extract the marginal distribution over the structural states:
\begin{align}
\ve{x}_{k\mid k-1}^\ast \sim \mathcal{N} \brc{ \ve{m}_{k \mid k-1}^{x,\ast}, \matr{P}_{k\mid k-1}^{x,\ast} }
\end{align}
from which we draw a representative sample denoted again as $\ve{x}_{k\mid k-1}^\ast$. This sample is passed through the trained BNN (from the mapping phase) to produce a predictive Gaussian distribution over the MFEs (cf. \cref{eq:pred_bnn_gaussian}). A sample $\tilde{\ve{\eta}}$ is then drawn from this distribution and treated as a \textit{pseudo-measurement}:
\begin{align} \label{eq:pseudogen}
\tilde{\ve{\eta}}_k \sim \mathcal{N} \brc{ \bar{\ve{\mu}}^{\eta}\brc{\ve{x}_{k\mid k-1}^\ast}, \bar{\matr{\Sigma}}^{\eta} \brc{\ve{x}_{k\mid k-1}^\ast } }
\end{align}
To relate this pseudo-measurement back to the augmented state vector, we define a measurement equation:
\begin{align} \label{eq:pseudo-meas_eq}
\tilde{\ve{\eta}}_k = \tilde{\matr{H}} \ve{z}_k^\ast + \tilde{\ve{v}}_k, \quad \tilde{\ve{v}}_k \sim \mathcal{N}(\ve{0}, \tilde{\matr{R}}_k),
\end{align}
where the measurement matrix $\tilde{\matr{H}} \in \mathbb{R}^{n \times 3n} = \begin{bmatrix} \matr{0}_{n \times 2n} & \mathbb{I}_{n \times n} \end{bmatrix}$ extracts the last $n$ components (corresponding to the LFs) from the $3n$-dimensional augmented state vector. The pseudo-measurement noise covariance is taken directly from the predictive posterior of  the BNN:
\begin{align} \label{eq:def_Rprog}
\tilde{\matr{R}}_k = \bar{\matr{\Sigma}}^{\eta}\brc{\ve{x}_{k\mid k-1}^\ast}
\end{align}

This formulation allows us to use the BNN output as probabilistic observations for assimilation, thereby closing the loop between learned nonlinear correction and linear Bayesian inference.

\subsubsection{Kalman update with BNN-based pseudo-measurements}
With the pseudo-measurement $\tilde{\ve{\eta}}_k$ and corresponding uncertainty $\tilde{\matr{R}}_k$ at hand, the Kalman update step refines the predicted distribution over the augmented state vector. This is achieved using the standard linear Kalman update equations:
\begin{subequations} \label{eq:prog_kal_update}
\begin{align}
   \matr{K}_k &= \matr{P}_{k\mid k-1}^{z,\ast} \tilde{\matr{H}}^T \brc{ \tilde{\matr{H}} \matr{P}_{k\mid k-1}^{z,\ast} \tilde{\matr{H}}^T + \tilde{\matr{R}}_k}^{-1}, \\
    \ve{m}_{k\mid k}^{z,\ast} &= \ve{m}_{k\mid k-1}^{z,\ast} + \matr{K}_k \brc{ \tilde{\ve{\eta}}_k - \tilde{\matr{H}} \ve{m}_{k\mid k-1}^{z,\ast}}, \\
    \matr{P}_{k\mid k}^{z,\ast} &= \brc{ \mathbb{I} - \matr{K}_k \tilde{\matr{H}} } \matr{P}_{k\mid k-1}^{z,\ast} 
\end{align}
\end{subequations}
Here, $\matr{K}_k$ is the Kalman gain matrix, $\ve{m}_{k\mid k}^{z,\ast}$ is the updated mean of the augmented state, and $\matr{P}_{k\mid k}^{z,\ast}$ is the updated covariance.

This update step effectively refines the distribution over both the structural states and the latent forces by assimilating pseudo-observations of MFEs derived from the BNN. As this process is repeated at each time step, it produces a probabilistic trajectory of the system response under the new excitation.

Moreover, this recursive mechanism naturally supports the online optimization of GP kernel hyperparameters for the prognosis-phase latent force model, which we address next.

\subsubsection{GP hyperparameter optimization}
To ensure accurate and stable state estimation under the new excitation $\ve{u}^\ast(t)$, we re-estimate the GP hyperparameters $\vtheta^\ast = \cbk{ \ell^\ast_j, \alpha^\ast_j }_{j=1}^n$ associated with the latent force model. Unlike the diagnosis phase, where these hyperparameters were learned from real physical measurements, here they are optimized using pseudo-measurements $\tilde{\ve{\eta}}_{1:N_t}$ generated via the BNN-conditioned mapping $p(\ve{\eta}^\ast \mid \ve{x}^\ast)$.

Once again, we adopt a MAP estimation strategy for this calibration, optimizing the following objective:
\begin{align} \label{eq:pred_gp_map}
\vtheta^\ast_{\text{MAP}} = \arg\max_{\vtheta^\ast} \cbk{ \log p \brc{ \tilde{\ve{\eta}}_{1:N_t^
\ast} \mid \vtheta^\ast } + \log p(\vtheta^\ast) }
\end{align}
Here, $p\brc{\tilde{\ve{\eta}}_{1:N_t^\ast} \mid \vtheta^\ast}$ is the pseudo-likelihood of the MFEs obtained from the Kalman filter predictions, and $p(\vtheta^\ast)$ is a prior distribution on the hyperparameters, chosen as a Student’s-$t$ prior to provide heavy-tailed regularization.

The likelihood term is evaluated using the pseudo-measurements as realizations from the BNN over MFEs given realizations of structural states $\ve{x}^\ast_k$ from Kalman filter prediction. This formulation maintains consistency with the estimation-stage optimization (\cref{subsec: optim Est.}) while adapting to the prognosis setting without requiring access to real measurements.

\subsubsection{Final state estimation and smoothing}
Once the GP hyperparameters $\vtheta^\ast_{\text{MAP}}$ have been optimized, the prognosis proceeds via a forward Kalman filtering pass followed by backward RTS smoothing. This two-stage procedure refines the structural state trajectories at each time step.

The final smoothed distribution over the augmented state vector is given by:
\begin{align}
\ve{z}_{k \mid N_t^\ast}^\ast \sim \mathcal{N} \brc{\ve{m}_{k \mid N_t^\ast}^{z,\ast}, \matr{P}_{k \mid N_t^\ast}^{z,\ast} }, \quad \text{for } k = 1, \ldots, N_t^\ast
\end{align}
from which the marginal distribution over the structural state $\ve{x}_k^\ast$ can be readily extracted. This smoothing procedure completes the inference pipeline for prognosis and yields the target predictive distribution:
\begin{align}
    p\brc{ \ve{x}_{1:N_t^\ast}^\ast \mid \ve{u}^\ast_{1:N_t^\ast}, \mathcal{D}_{\text{obs}}, \ve{\phi} } \equiv p\brc{ \ve{x}_{1:N_t^\ast}^\ast \mid \ve{u}^\ast_{1:N_t^\ast}, \tilde{\ve{\eta}}^\ast_{1:N_t^\ast}, \ve{\phi} }
\end{align}
This equivalence holds because the diagnostic dataset $\mathcal{D}_{\text{obs}}$ is implicitly encoded in the pseudo-measurements $\tilde{\ve{\eta}}_{1:N_t^\ast}$ through the learned conditional mapping $p(\ve{\eta} \mid \ve{x})$. This mapping was trained using samples from the joint posterior distribution $p \brc{\ve{x}_k, \ve{\eta}_k \mid \mathcal{D}_{\text{obs}}, \ve{\phi}}$ obtained during the diagnosis phase, ensuring that the prognosis remains informed by prior system behavior.

\begin{algorithm}
\caption{\textbf{Prognosis}: State Prediction under New Input Using Pseudo-Measurements}\label{algo: Prognosis}
\begin{algorithmic}[1]
\Statex \textbf{Input:} New external input signal $\ve{u}^\ast_{1:N_t^\ast}$, initial state $\ve{x}_0^\ast$, structural parameters $\ve{\phi}$, trained BNN $\pr{\ve{\eta} \mid \ve{x}}$
\Statex \textbf{Output:} Prognostic posterior $\pr{\ve{x}_k^\ast \mid \ve{u}^\ast_{1:N_t^\ast}, \tilde{\ve{\eta}}_{1:N_t^\ast}, \ve{\phi}}$
\vspace{0.3em}
\Statex \textbf{Step 1: Prognostic model setup}
\State Model the MFE $\ve{\eta}^\ast(t)$ as a stationary GP with hyperparameters $\vtheta^\ast$
\State Construct the augmented state vector $\ve{z}^\ast = \sbk{ \brc{\ve{x}^\ast}^T \;\; \brc{\ve{\eta}^\ast}^T }^T$
\State Construct the discrete-time augmented process model as given by \cref{eq:Process_eq_pred_disc}
\vspace{0.3em}
\Statex \textbf{Step 2: GP hyperparameter estimation}
\State Optimize GP hyperparameters $\vtheta^\ast$ via MAP estimation using pseudo-measurements generated by BNN (\cref{eq:pred_gp_map}), which involves computing the likelihood using the following loop
\For{$k=1$ to $N_t^\ast$}
    \State Predict mean $\ve{m}_{k\mid k-1}^{x,\ast}$ and covariance $\matr{P}_{k\mid k-1}^{x,\ast}$ using \cref{eq: Prog-KF pred}
    \State Generate a sample of structural state prediction $\ve{x}_{k\mid k-1}^\ast \sim \mathcal{N}\brc{\ve{m}_{k\mid k-1}^{x,\ast}, \matr{P}_{k\mid k-1}^{x,\ast}}$ 
    \State Generate pseudo-measurement $\tilde{\ve{\eta}}_k$ using \cref{eq:pseudogen}
    \State Define measurement matrix $\tilde{\matr{H}}$, and set the measurement noise covariance $\tilde{\matr{R}}_k$ using \cref{eq:def_Rprog}
    \State Update the state estimate using \cref{eq:prog_kal_update}
\EndFor
\State Post GP parameter optimization, perform Kalman filtering and RTS smoothing to obtain final state estimates $\pr{\ve{x}_{1:N_t^\ast} \mid \ve{u}_{1:N_t^\ast}^\ast, \tilde{\ve{\eta}}_{1:N_t^\ast}, \ve{\phi}}$ 

\end{algorithmic}
\end{algorithm}

\section{Numerical studies} \label{sec: Numerical studies}

We now present a set of numerical and benchmark case studies to evaluate the proposed diagnosis-to-prognosis framework. The examples span systems with different types of model-form errors (e.g., static and dynamic nonlinearities), varying dimensionality (from SDOF to multi-DOF), and include both simulated and experimental data. For each case, we implement the full pipeline comprising: (1) \textbf{Diagnosis} via GPLFM-based joint estimation of system states and latent MFEs; (2) \textbf{Mapping} via training a Bayesian neural network on posterior samples to capture the probabilistic relationship between system states and latent forces; and (3) \textbf{Prognosis} wherein state predictions is performed for unseen (test) input excitations in Kalman filtering using GPLFM by incorporating pseudo-measurements generated from the learned probabilistic mapping, in absence of actual measurement data.

The first two numerical studies involve simulated systems excited with a non-stationary ground acceleration input during the diagnosis phase. The input is generated by passing Gaussian white noise through a Kanai–Tajimi filter \cite{kanai_taj}, modulated by a non-stationary envelope to mimic earthquake-like ground motion. This excitation is applied over a duration of 60 seconds and sampled at 200 Hz. System states are initialized to zero. Output measurements are taken as accelerations, corrupted with Gaussian noise of zero mean and standard deviation equal to $5\%$ of the RMS of the true signal.

\begin{figure}[H]
    \centering
    \includegraphics[scale=0.55]{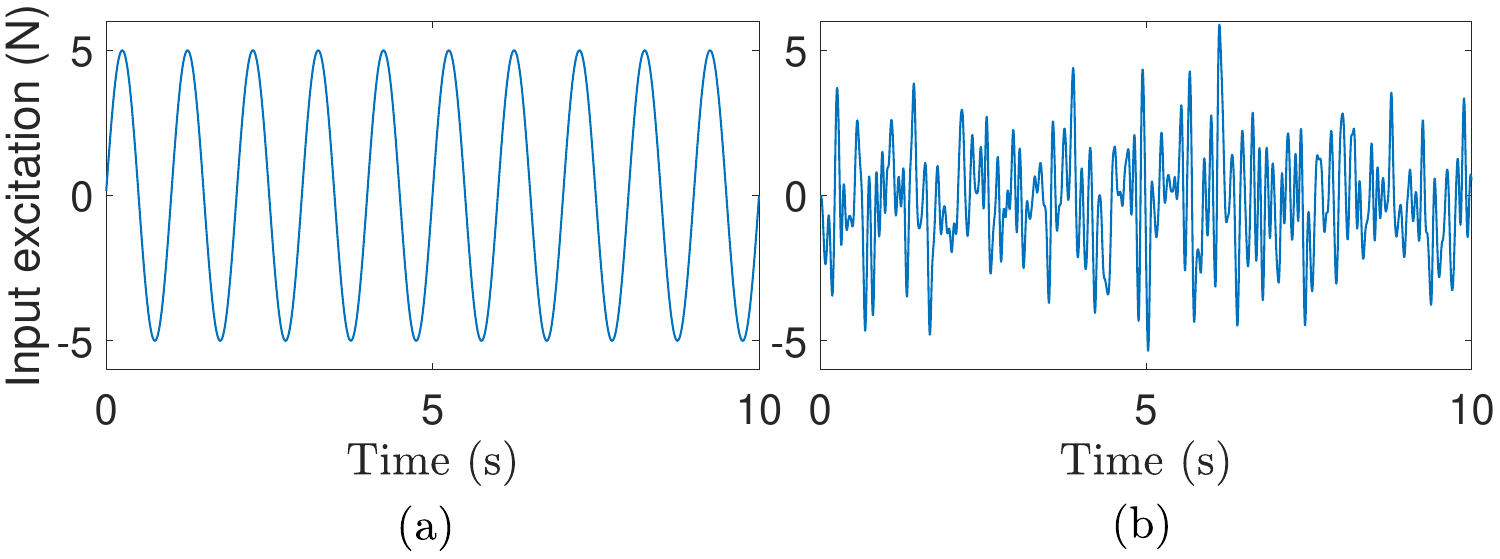}
    \caption{External force input signals used during the prognosis phase to evaluate model generalization. (a) Periodic sinusoidal input representing structured forcing. (b) Filtered broadband noise representing stochastic excitation.}
   \label{fig:combined_input}
\end{figure}

For prognosis, two types of unseen external force excitations are used: (a) sinusoidal signal, and (b) filtered white noise via a low-pass Butterworth filter. These are shown in \Cref{fig:combined_input}. In the SDOF case, the external force acts at the only available DOF, while in the 3-DOF system, prognosis is performed using force excitations at different DOFs than during diagnosis, to demonstrate spatial generalizability.
The third and fourth examples correspond to benchmark systems: the Silverbox system with a static cubic nonlinearity, and the Bouc–Wen hysteretic system with history-dependent dynamics. Detailed configurations of these systems are provided in their respective subsections.

In all examples, diagnosis is performed using a Kalman filter with the known physics and GP model for the MFEs, while mapping involves training a BNN on ensembles sampled from the smoothed joint posterior from the diagnosis phase. The BNN consists of two hidden layers with 20 and 10 neurons and ReLU activation unless mentioned otherwise. Posterior predictive distributions from the BNN are used to generate pseudo-measurements in the prognosis phase. For both diagnosis and prognosis phase, the GP hyperparameters are optimised using MATLAB’s \cite{MATLAB:R2023b_u9} \texttt{fmincon} with the interior-point algorithm and positive bounds $[10^{-15}, 10^{15}]$.

True structural states for comparison are obtained via a fourth-order Runge–Kutta integration of the ground-truth governing equations.

\subsection{SDOF Duffing oscillator} \label{subsec: Toy Duffing}
\begin{figure}[H]
    \centering
    \includegraphics[scale = 1]{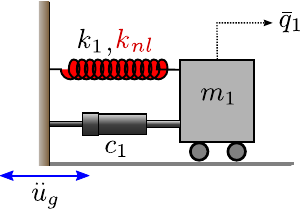}
    \caption{Schematic of the SDOF oscillator with Duffing-type nonlinear stiffness, excited by ground acceleration $\ddot{u}_g$ during diagnosis.}
    \label{fig:true_SDOF_system}
\end{figure}

We consider an SDOF mass-spring-damper oscillator with a cubic nonlinear stiffness element, as depicted in \Cref{fig:true_SDOF_system}. 
The true dynamics of the oscillator is governed by the nonlinear equation of motion:
\begin{align}
    m \ddot{\bar{q}}(t) + c \dot{\bar{q}}(t) + k \bar{q}(t) + k_{\text{nl}} \bar{q}^3(t) = -m \ddot{u}_g(t)    
\end{align}
where the nonlinear restoring force is given by $p(\bar{q}(t)) = k_{\text{nl}} \bar{q}^3(t)$, with $\bar{q}(t)$ denoting the relative displacement. The nominal model assumes a linear form, neglecting the nonlinear restoring force. The MFE arising from this discrepancy is modeled as a latent force $\eta(t)$ and incorporated into the corrected dynamics as described in \cref{eq:corrected_mdof_EoM}. The physical parameters are listed in \cref{tab:SDOF_parameter}.

\begin{table}[h!]
\centering
\begin{tabular}{|c|c|c|c|c|}
\hline
\textbf{Parameter} & $m$ & $k$ & $c$ & ${k}_\text{nl}$ \\ \hline
\textbf{Value} & $1\, \si{kg}$ & $100\, \si{N/m}$ & $0.2\, \si{Ns/m}$ & $1000\, \si{N/m^3}$ \\ \hline
\end{tabular}
\caption{System parameters for the SDOF Duffing oscillator.}
\label{tab:SDOF_parameter}
\end{table}

\begin{figure}[H]
	\centering
	\includegraphics[width=\textwidth]{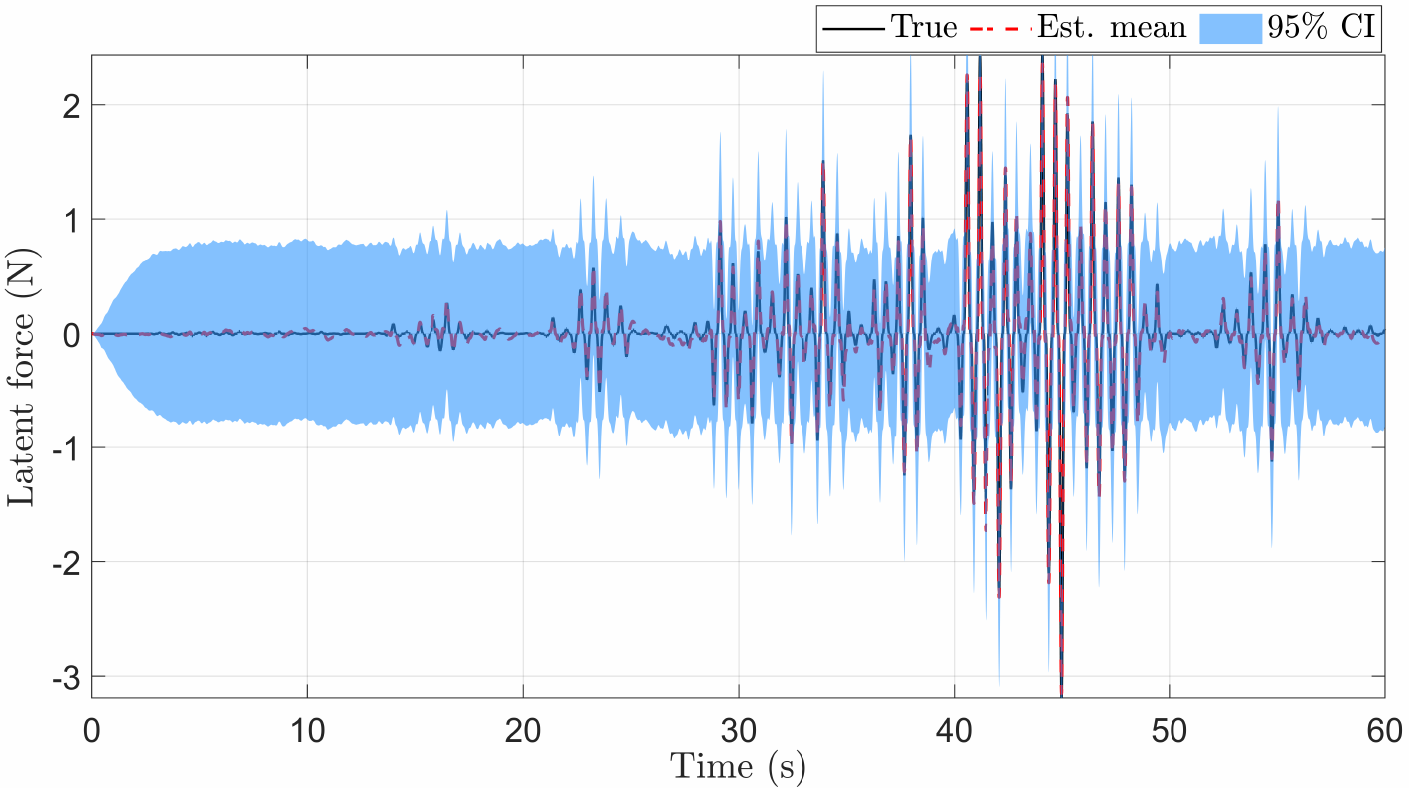}
	    \caption{\textbf{Diagnosis}: Estimated latent force $\eta(t)$ versus true nonlinear restoring force. The shaded region indicates $\pm2\sigma$ confidence interval.}
        \label{fig: SDOF_diagnosis_MFE}
\end{figure}

\begin{figure}[H]
	\centering
	\includegraphics[width=\textwidth]{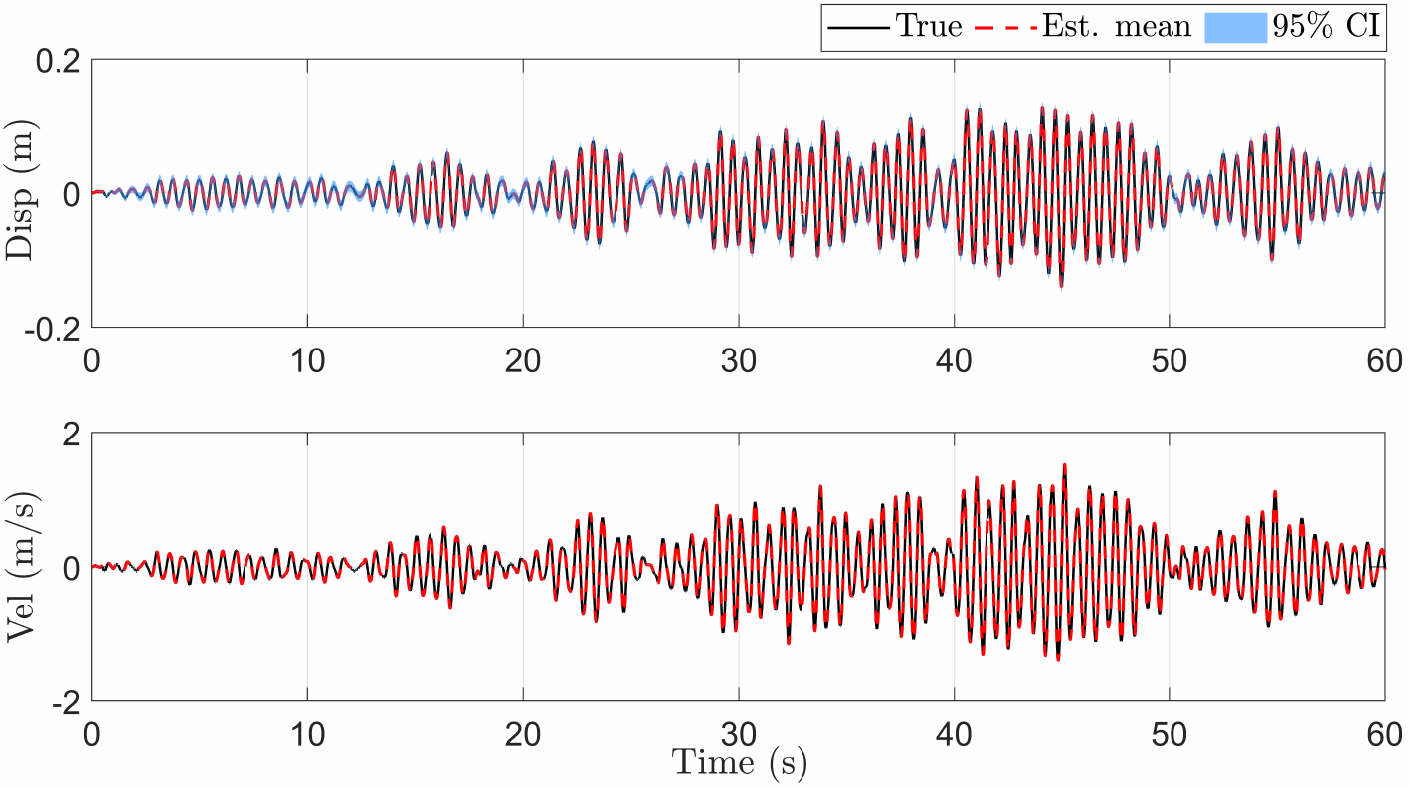}
	    \caption{\textbf{Diagnosis}: Estimated system states (displacement and velocity) with $\pm2\sigma$ confidence bounds versus true states.}
        \label{fig: SDOF_diagnosis_state}
\end{figure}

The diagnosis phase aims to estimate the full states $\ve{x}(t)$ and latent force $\eta(t)$ from noisy acceleration measurements obtained under ground excitation. Within the GPLFM framework, the available data are assimilated, and the GP hyperparameters are optimized to yield the smoothed joint posterior distribution over the system states $\ve{x}_k = [q_k\;\; \dot{q}_k]^T$ and the latent force $\eta_k$, for $k=1,\dots,N_t$. As shown in \Cref{fig: SDOF_diagnosis_MFE}, the estimated mean of $\eta(t)$ closely follows the true nonlinear restoring force $p(\bar{q}(t))$, albeit with relatively wide confidence intervals, indicative of higher uncertainty in MFE inference. In contrast, \Cref{fig: SDOF_diagnosis_state} demonstrates excellent agreement between the estimated mean and true states, with significantly narrower confidence bounds, reflecting more precise state estimation.

\begin{figure}[H]
	\centering
	\includegraphics[width=\textwidth]{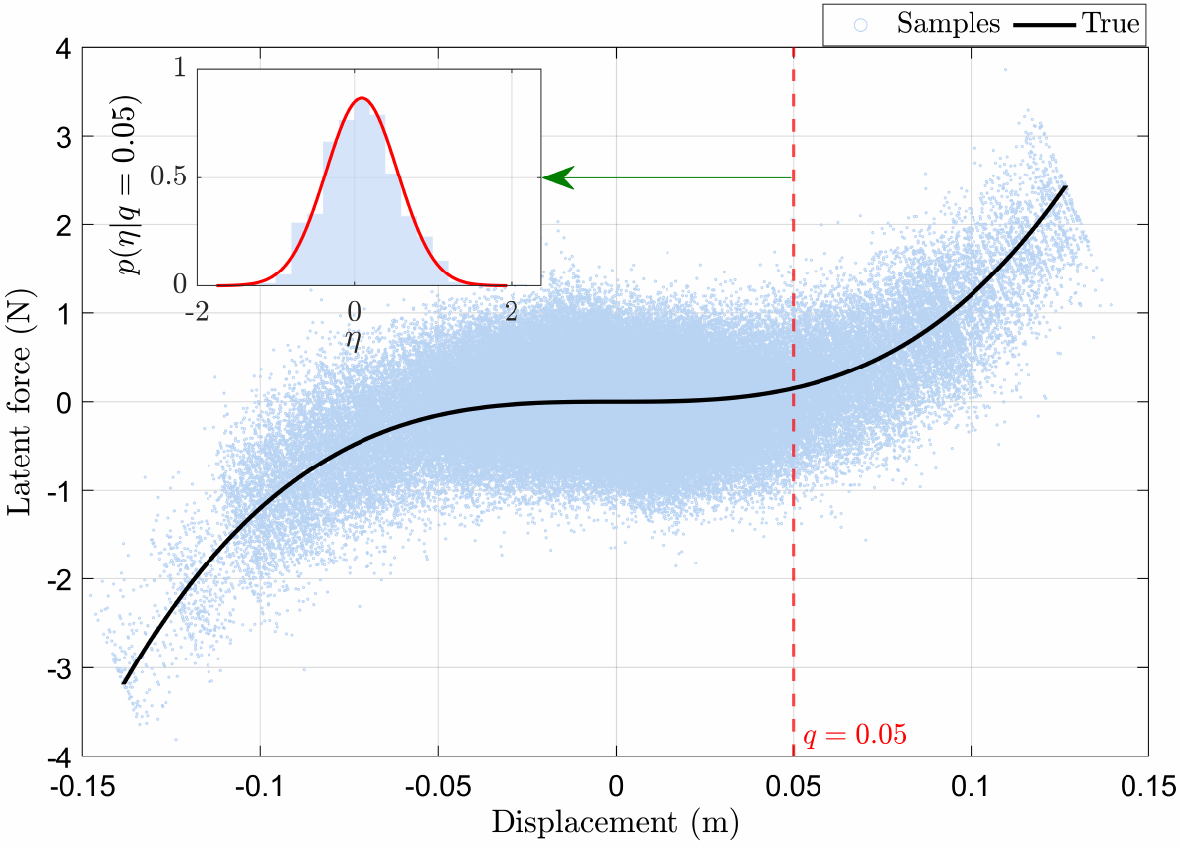}
        \caption{Samples of displacement $q$ versus latent force $\eta$ used to train the BNN. The inset shows a conditional Gaussian fit at $q=0.05$, highlighting input-dependent variance.}
        \label{fig: ensemble_x_vs_eta}
\end{figure}

Following diagnosis, the mapping phase trains a BNN to learn the conditional distribution $p(\eta \mid \ve{x})$ using samples from the smoothed posterior. The BNN models both the mean and the variance of $\eta$ as functions of $\ve{x}$, capturing heteroskedasticity in the data. \Cref{fig: ensemble_x_vs_eta} shows the sample cloud of displacement $q$ versus $\eta$, with an inset Gaussian fit at $q = 0.05$ illustrating the conditional distribution.

In the prognosis phase, the learned mapping is incorporated into a Kalman filtering framework via pseudo-measurements, enabling forward state prediction under new external force excitations.

\subsubsection*{Prognosis under sinusoidal excitation}
\begin{figure}[H]
	\centering
	\includegraphics[width=\textwidth]{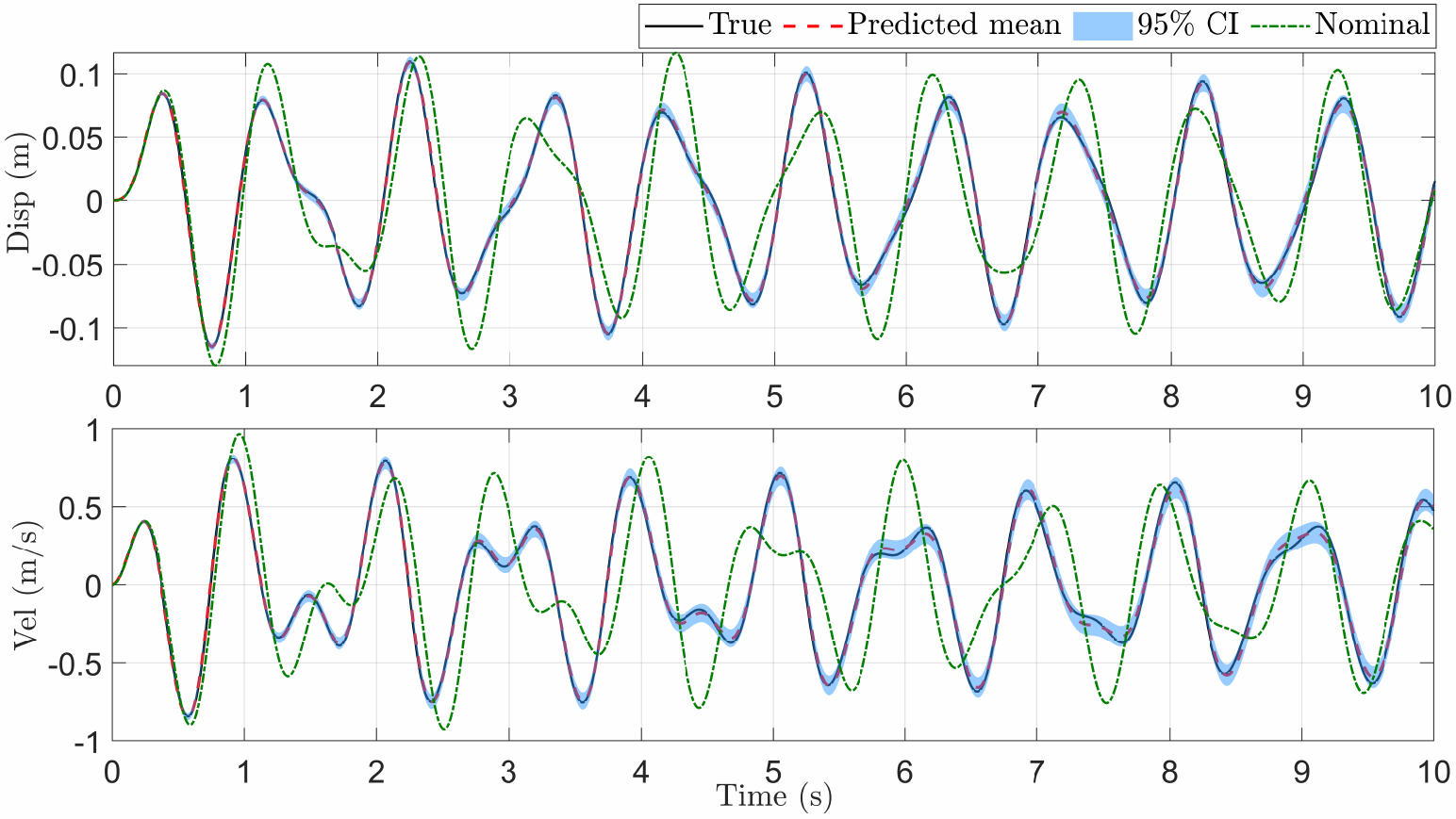}
	    \caption{\textbf{Prognosis}: Predicted system states under sinusoidal excitation using the GPLFM framework. The $\pm2\sigma$ confidence interval captures the true response. Nominal model prediction shown for comparison.}
        \label{fig: SDOF_pred_sin_input}
\end{figure}

The system response to a sinusoidal input (\Cref{fig:combined_input}) is predicted using the proposed framework as shown in \textcolor{black}{\Cref{fig: SDOF_pred_sin_input}}. The predicted state distribution closely matches the true trajectory, with uncertainty bounds enveloping the ground truth. In contrast, the misspecified nominal linear model exhibits significant deviation due to unaccounted nonlinearities, underscoring the importance of MFE diagnosis.

\subsubsection*{Prognosis under filtered white noise excitation}
\begin{figure}[H]
	\centering
	\includegraphics[width=\textwidth]{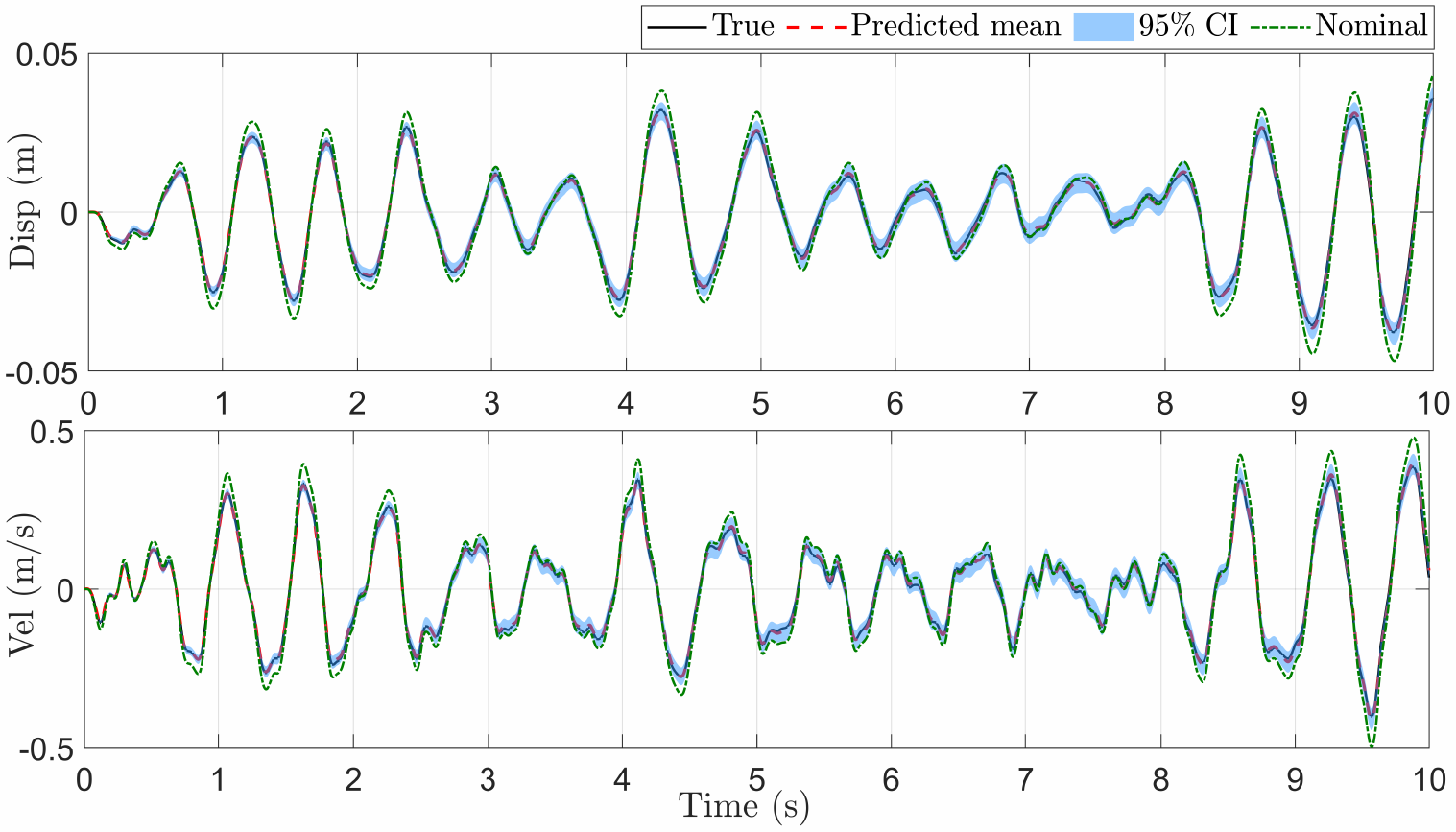}
	\caption{\textcolor{black}{\textbf{Prognosis}: Predicted system states under filtered white noise excitation. The GPLFM framework closely tracks the true states, with $\pm2\sigma$ confidence bounds quantifying uncertainty. Nominal model prediction shown for comparison.}} 
        \label{fig: SDOF_white_noise_input}
\end{figure}

Under stochastic excitation from filtered white noise (cf.\ \Cref{fig:combined_input}), the framework continues to deliver accurate predictions as shown in \textcolor{black}{\Cref{fig: SDOF_white_noise_input}}. The predicted means align well with the true system states, and the associated confidence intervals effectively quantify the predictive uncertainty. \textcolor{black}{For reference, similar to the sinusoidal excitation case, the predictions from the nominal model are also included in \Cref{fig: SDOF_white_noise_input}. 
As expected, the nominal model---owing to its incomplete physics---exhibits poor predictive
performance compared to the LF-corrected model, further emphasizing the importance of the proposed methodology.} 

\subsection{3-DOF system with local nonlinearities}
\begin{figure}[H]
    \centering
    \includegraphics[scale = 1]{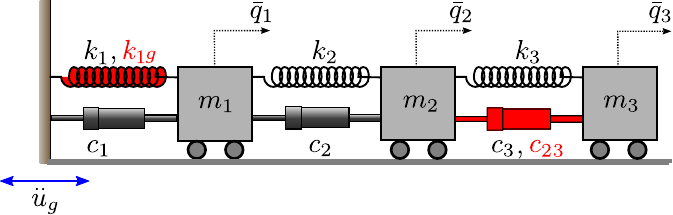}
    \caption{Schematic of the 3-DOF system with cubic spring stiffness between ground and mass 1 and quadratic damping between masses 2 and 3.}
    \label{fig:true_3DOF_system}
\end{figure}

To assess the proposed framework's performance on an MDOF system, we consider a 3-DOF shear-building model featuring localized nonlinearities, as depicted in \Cref{fig:true_3DOF_system}. 
Each floor has mass $m_i = \SI{1}{kg}$, stiffness $k_i = \SI{100}{N/m}$, and damping $c_i = \SI{0.2}{Ns/m}$ for $i = 1,2,3$. Two nonlinear effects are introduced: a cubic stiffness (Duffing-type) spring with coefficient $k_{1g} = \SI{1000}{N/m^3}$ between the ground and DOF 1, and a quadratic damping element with coefficient $c_{23} = \SI{0.5}{Ns^2/m^2}$ between DOFs 2 and 3. These unmodeled physics contribute to MFEs, represented as latent forces $\ve{\eta}(t)$:
\begin{align}
    \ve{p}\brc{\bar{\ve{q}}(t),\dot{\bar{\ve{q}}}(t)} =
    \begin{Bmatrix}
        k_{1g}\bar{q}_1^3(t) \\
        0 \\
        c_{23}\left(\dot{\bar{q}}_3(t) - \dot{\bar{q}}_2(t)\right)\left|\dot{\bar{q}}_3(t) - \dot{\bar{q}}_2(t)\right|
    \end{Bmatrix}
\end{align}

\begin{figure}[H]
	\centering
	\includegraphics[width=\textwidth]{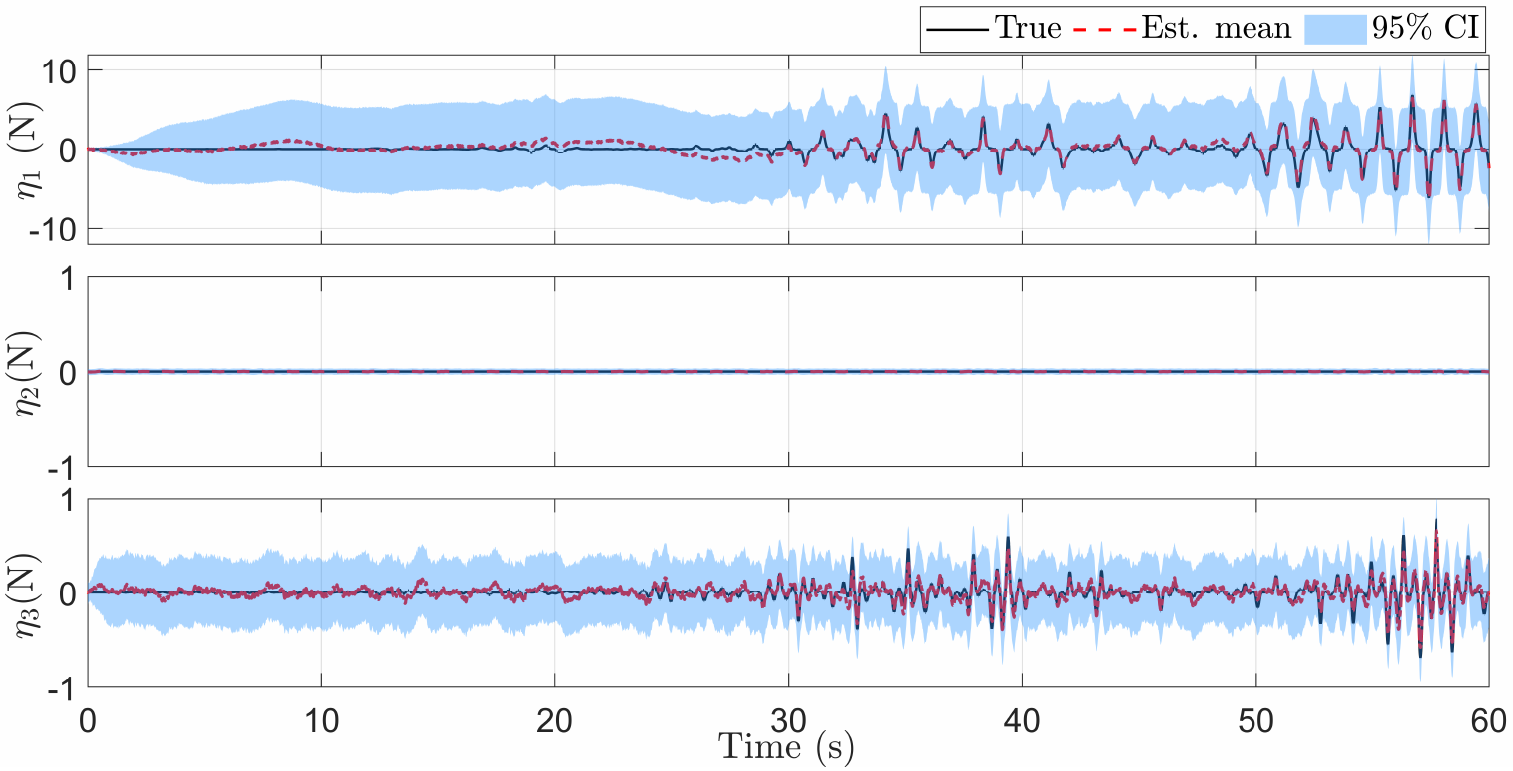}
        \caption{\textbf{Diagnosis}: Estimated latent forces $\ve{\eta}$ versus true nonlinear forces at each DOF of the 3-DOF system with $\pm2\sigma$ confidence intervals.}
        \label{fig: 3DOF_diagnosis_MFE}
\end{figure}

\begin{figure}[H]
	\centering
	\includegraphics[width=\textwidth]{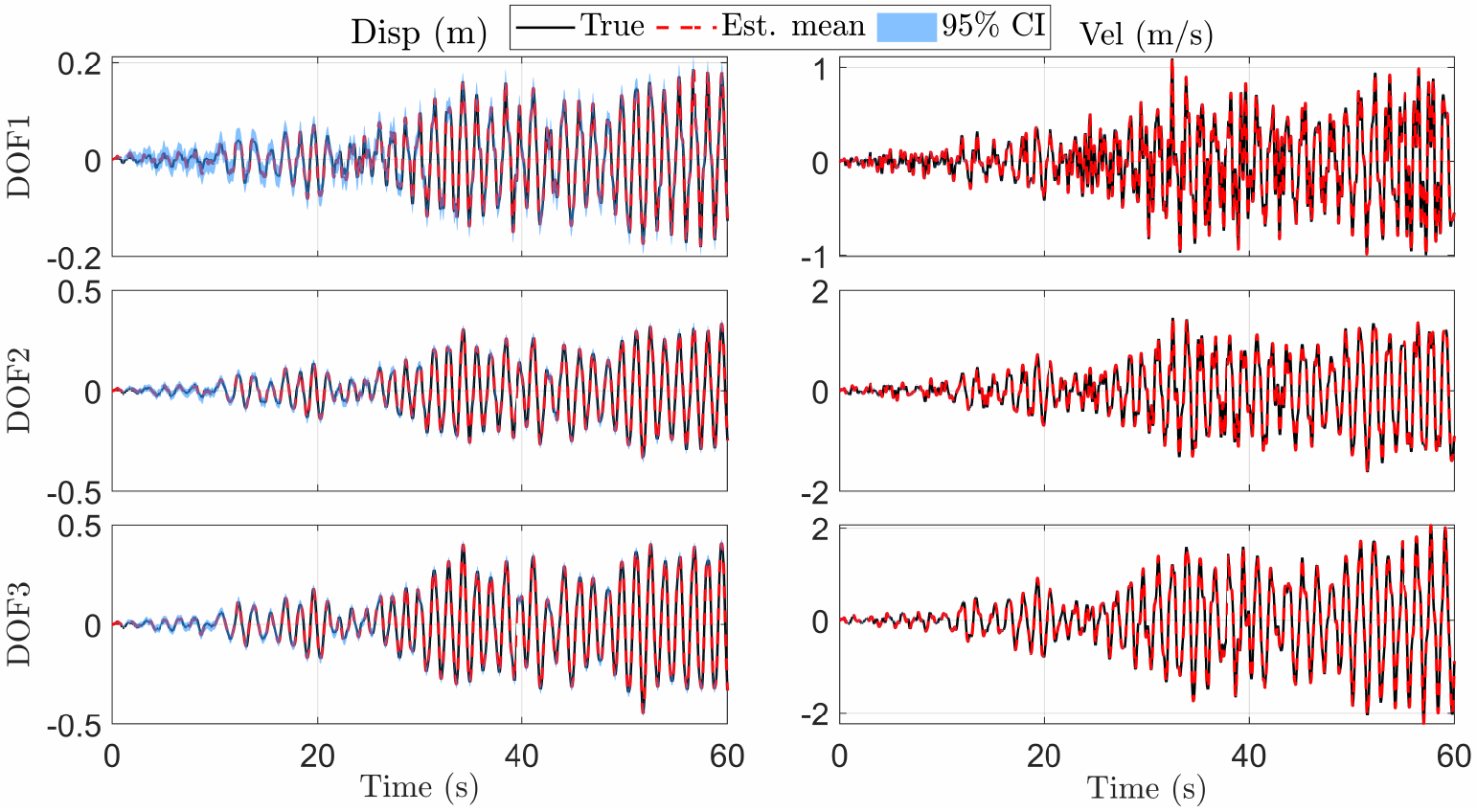}
	    \caption{\textbf{Diagnosis}: Estimated displacements and velocities versus true states at all DOFs of the 3-DOF system, with $\pm2\sigma$ confidence intervals.}
        \label{fig: 3DOF_diagnosis_state}
\end{figure}

The diagnosis phase estimates $\ve{\eta}(t)$ and system states from noisy acceleration measurements under ground excitation. \Cref{fig: 3DOF_diagnosis_MFE} shows the inferred mean latent forces aligning well with their true counterparts, while \Cref{fig: 3DOF_diagnosis_state} presents the estimated states. DOFs 1 and 3 show non-zero MFEs due to the localized nonlinearities, whereas DOF 2 exhibits negligible MFE, with posterior mean and confidence intervals tightly concentrated around zero. This sparsity arises from the Student's-$t$ prior used in GP hyperparameter learning, which penalizes unnecessary complexity.


Posterior samples of states and latent forces are then used in the mapping phase to train a BNN that learns $p(\ve{\eta} \mid \ve{x})$, where $\ve{x}$ denotes the full structural system state vector comprising displacements and velocities at all DOFs. This full-state input formulation is deliberately chosen to maintain a general, flexible mapping without any a priori assumptions about the sparsity or locality of the MFEs.

However, in the context of the present example, such structural insights are indeed available. Specifically, the MFE at DOF 1 arises solely due to a nonlinear spring between the ground and mass 1, and hence depends only on the displacement at DOF 1. Similarly, the MFE at DOF 3 originates from a nonlinear damping element between masses 2 and 3, and is a function of the relative velocity between DOFs 2 and 3. These facts suggest that more compact, physically-informed mappings --- using only the relevant subset of state variables for each DOF --- could potentially improve efficiency and interpretability. Nevertheless, in this work, we refrain from exploiting such structural prior knowledge in the mapping phase. This choice ensures generality and paves the way for application to more complex systems where such a priori knowledge may be unavailable or uncertain.

\subsubsection*{Prognosis under sinusoidal excitation}
To quantitatively assess the predictive performance of the framework, we use the normalized mean square error (NMSE), defined for a generic vector time-series signal $\ve{d}(t_k)$, having $n$ components and $N_t^\ast$ time points, as:
\begin{align}
    {\text{NMSE}} = \left[ \frac{1}{n N_t^\ast} \sum_{i=1}^{n} \sum_{k=1}^{N_t^\ast} \frac{\left( d_i(t_k) - \hat{d}_i(t_k) \right)^2}{\sigma_i^2} \right] \times 100\%
\end{align}
where $d_i(t)$ denotes the true signal, $\hat{d}_i(t)$ is the predicted mean signal, and $\sigma_i^2$ is the variance of $d_i(t)$. This metric is applied separately to both displacement and velocity vector signals to compute the corresponding NMSE values.

\begin{figure}[H]
    \centering
    \includegraphics[width=\textwidth]{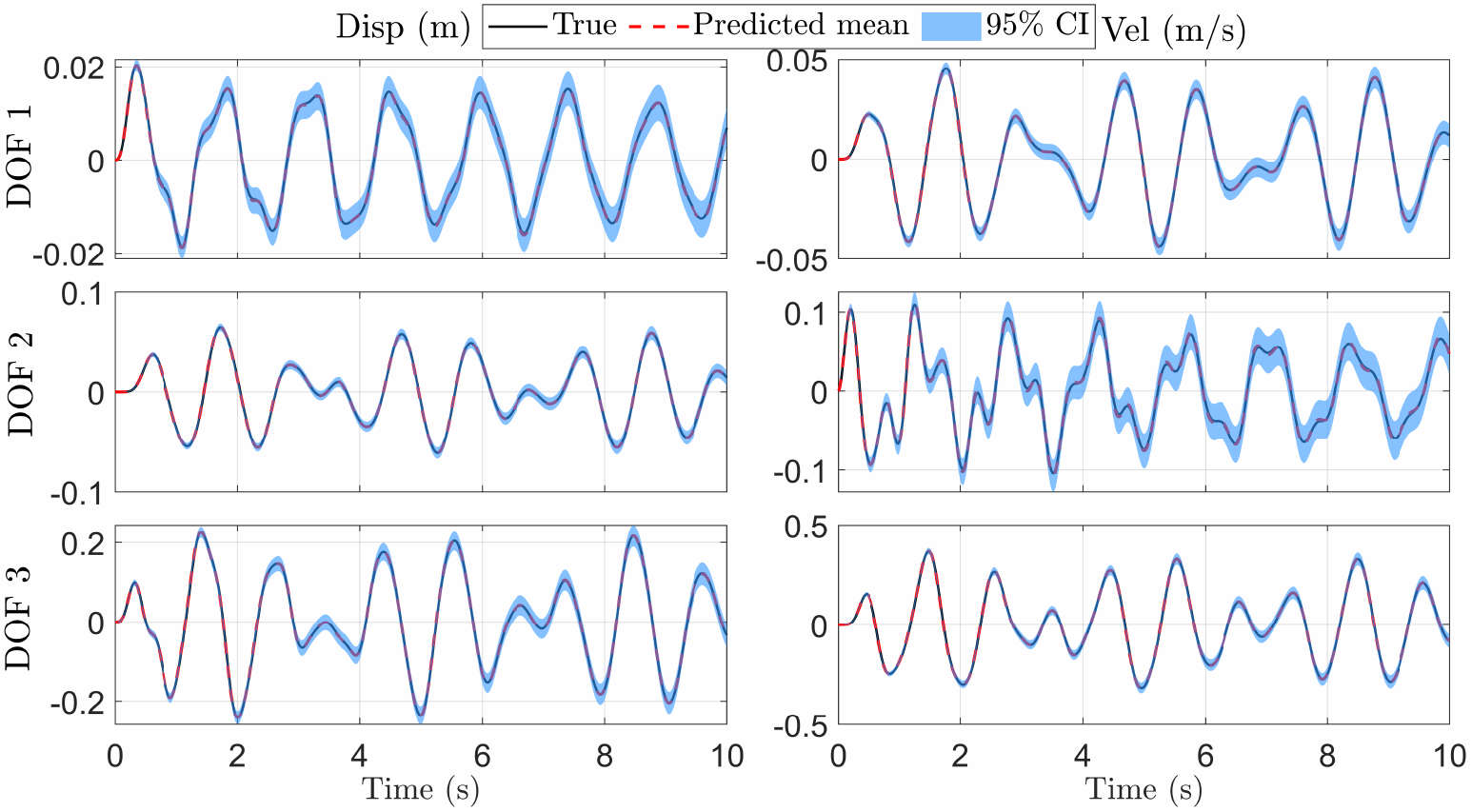}
    \caption{\textbf{Prognosis}: Predicted system states under sinusoidal excitation at DOF 1. The shaded region denotes $\pm2\sigma$ confidence bounds.}
    \label{fig: 3DOF_pred_sin_input_DOF1}
\end{figure}

Figure~\ref{fig: 3DOF_pred_sin_input_DOF1} shows the predicted system states when a sinusoidal excitation is applied at DOF 1. The mean predictions closely track the true system trajectories, with $2\sigma$ confidence bounds effectively enveloping the ground truth. The NMSEs in displacement and velocity vectors are $0.0309\%$ and $0.0595\%$, respectively. 

To systematically evaluate the framework's predictive performance, this process is repeated for two additional scenarios where the sinusoidal excitation is applied independently at DOFs 2 and 3. In each case, the NMSE is computed separately for the displacement and velocity vector signals. Across all three scenarios (DOF 1, 2, and 3 excitations), the NMSE values for both displacement and velocity remain below $1\%$, highlighting the robustness of the proposed framework in multi-DOF settings with localized nonlinearities.

\subsubsection*{Prognosis under filtered white noise excitation}
\begin{figure}[H]
    \centering
    \includegraphics[width=\textwidth]{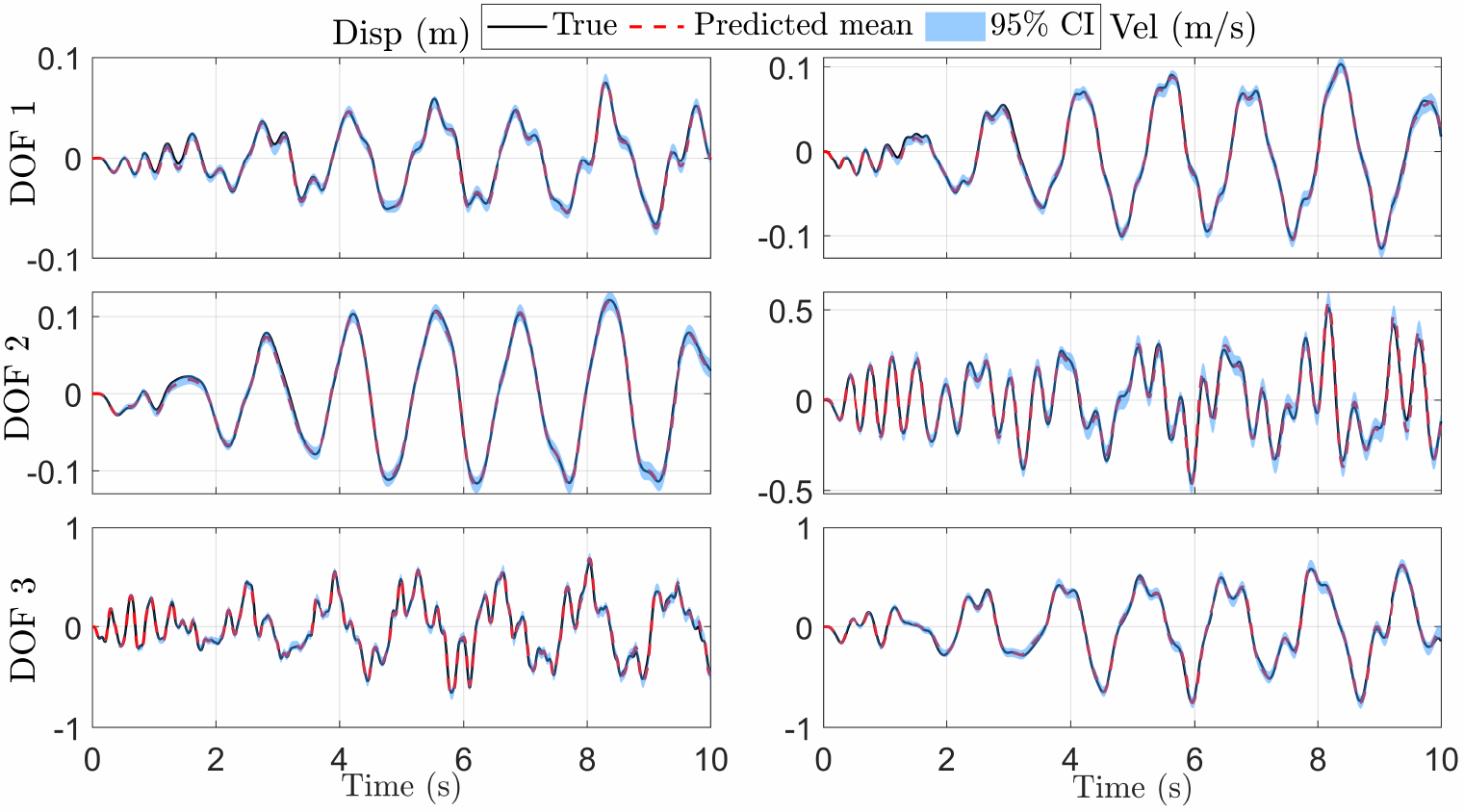}
    \caption{\textbf{Prognosis}: Predicted system states under filtered white noise excitation at DOF 1, with $\pm2\sigma$ confidence intervals.}
    \label{fig:3DOF_whiteNoise_input_DOF1}
\end{figure}

To further assess the robustness of the proposed prediction framework, we consider individual cases where the system is subjected to filtered white noise excitation applied independently at each DOF. We first present the state prediction results corresponding to excitation at DOF 1 in \Cref{fig:3DOF_whiteNoise_input_DOF1}, where the predicted displacements and velocities closely follow the true states, and the corresponding $2\sigma$ CIs reliably capture the ground truth trajectories. Using the NMSE metric, the errors for displacement and velocity are found to be $0.3527\%$ and $0.6849\%$, respectively. For excitations applied at DOFs 2 and 3, the NMSE values for both state variables remain below $3\%$.

Although slightly higher than those observed under sinusoidal input, due to the broader spectral content of filtered white noise, the still low NMSE values show the framework’s ability to generalize beyond the training conditions. These results highlight the framework’s resilience in capturing nonlinear model discrepancies and delivering accurate state predictions under diverse and uncertain excitation environments --- an essential requirement for real-world deployment in structural health monitoring or digital twin applications.

\subsection{Silverbox Benchmark} 
This benchmark study evaluates the proposed GPLFM-based framework on the Silverbox system --- an experimental SDOF electronic circuit that mimics the dynamics of a Duffing oscillator. Originally introduced as part of a nonlinear system identification challenge \cite{silverBox}, this system is widely adopted to test identification and prediction algorithms in nonlinear settings \cite{2022RogersNonlinearIdentification,nayek2021spike,SiverBox1,Silverbox2}. Although the Silverbox does not exactly conform to the theoretical Duffing oscillator equation, its response closely approximates such behavior, making it a relevant and challenging benchmark.

\begin{figure}[H]
    \centering
    \includegraphics[width=\textwidth]{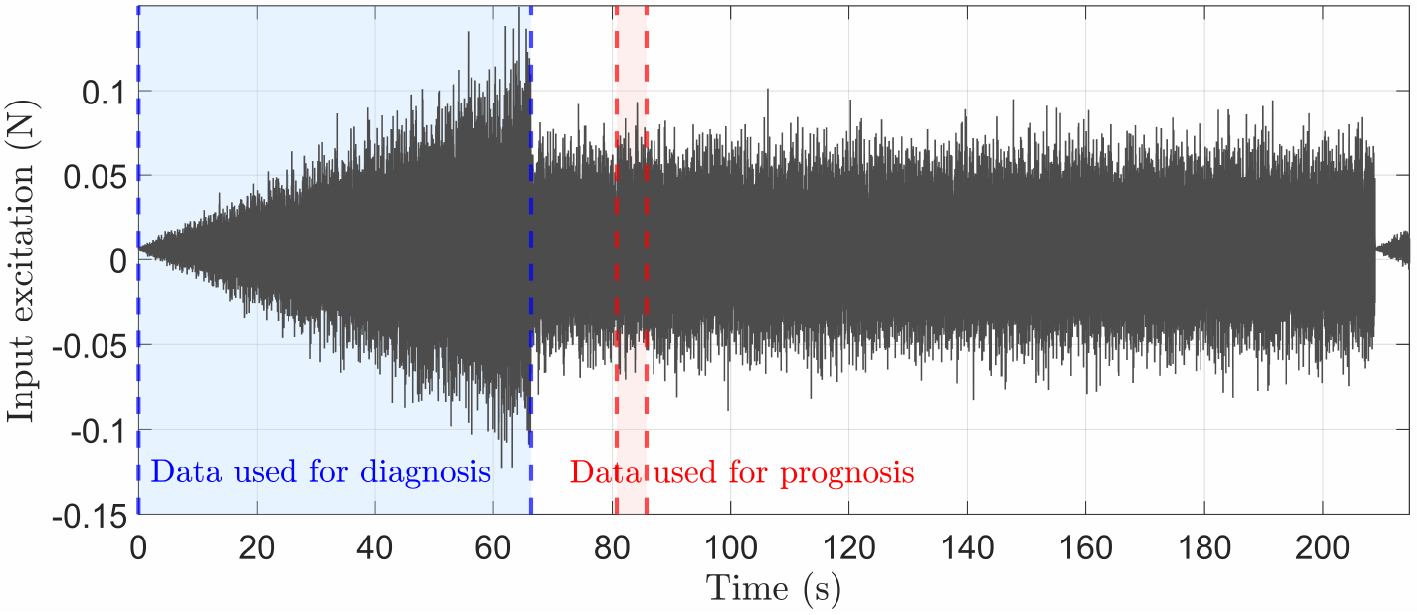}
    \caption{Input excitation signal from the Silverbox dataset. The segment highlighted in red is used for diagnosis; the segment in blue is used for prognosis.} 
    \label{fig: SilverBox_input}
\end{figure}

We focus on the \texttt{arrowhead} dataset, available in the file \texttt{SNLS80mV.mat}, which contains the input signal \texttt{V1} (interpreted as input force $u(t)$) and output signal \texttt{V2} (interpreted as displacement $q(t)$). As shown in \Cref{fig: SilverBox_input}, the input consists of two parts: a filtered Gaussian white noise excitation with increasing amplitude (first 40,000 samples) and ten realizations of an odd random phase multisine excitation (each with 8192 samples).

The diagnosis phase uses one realization of the multisine input (samples 49,278 to 52,350), with corresponding output displacement values. The prognosis phase uses a non-overlapping portion of the input signal --- the gradually increasing amplitude Gaussian excitation --- for prediction.

To simulate model-form error, we define the true nonlinear system dynamics using the Duffing-type equation:
\begin{align}\label{eq:SilverBox_true_EoM}
    m\ddot{\bar{q}}(t) + c\dot{\bar{q}}(t) + k\bar{q}(t) + k_\text{nl} \bar{q}^3(t) = u(t)
\end{align}
While the exact physical parameters are not known, we adopt the estimates from \cite{2022RogersNonlinearIdentification} listed in \Cref{tab:SilverBox_params}. The nominal model used in our framework is a linear version of \cref{eq:SilverBox_true_EoM}, where the nonlinear term is excluded. The MFE arising from this omission is absorbed as a latent force, modeled by a GP. This LF corresponds to the nonlinear restoring force: $\eta(t) \approx k_\text{nl} \bar{q}^3(t)$.

\begin{table}[h!]
    \centering
    \begin{tabular}{|c|c|c|c|c|}
        \hline
        \textbf{Parameter} & $m$  & $k$  & $c$  & $k_\text{nl}$  \\ \hline
        \textbf{Value} & $5.3722\times10^{-6}\, \si{kg}$ & $0.9932\, \si{N/m}$ & $2.1905\times10^{-4}\, \si{Ns/m}$ & $3.4239\, \si{N/m^3}$ \\ \hline
    \end{tabular}
    \caption{Silverbox system parameters (adopted from \cite{2022RogersNonlinearIdentification})}
    \label{tab:SilverBox_params}
\end{table}

\begin{figure}[H]
    \centering
    \includegraphics[width=\textwidth]{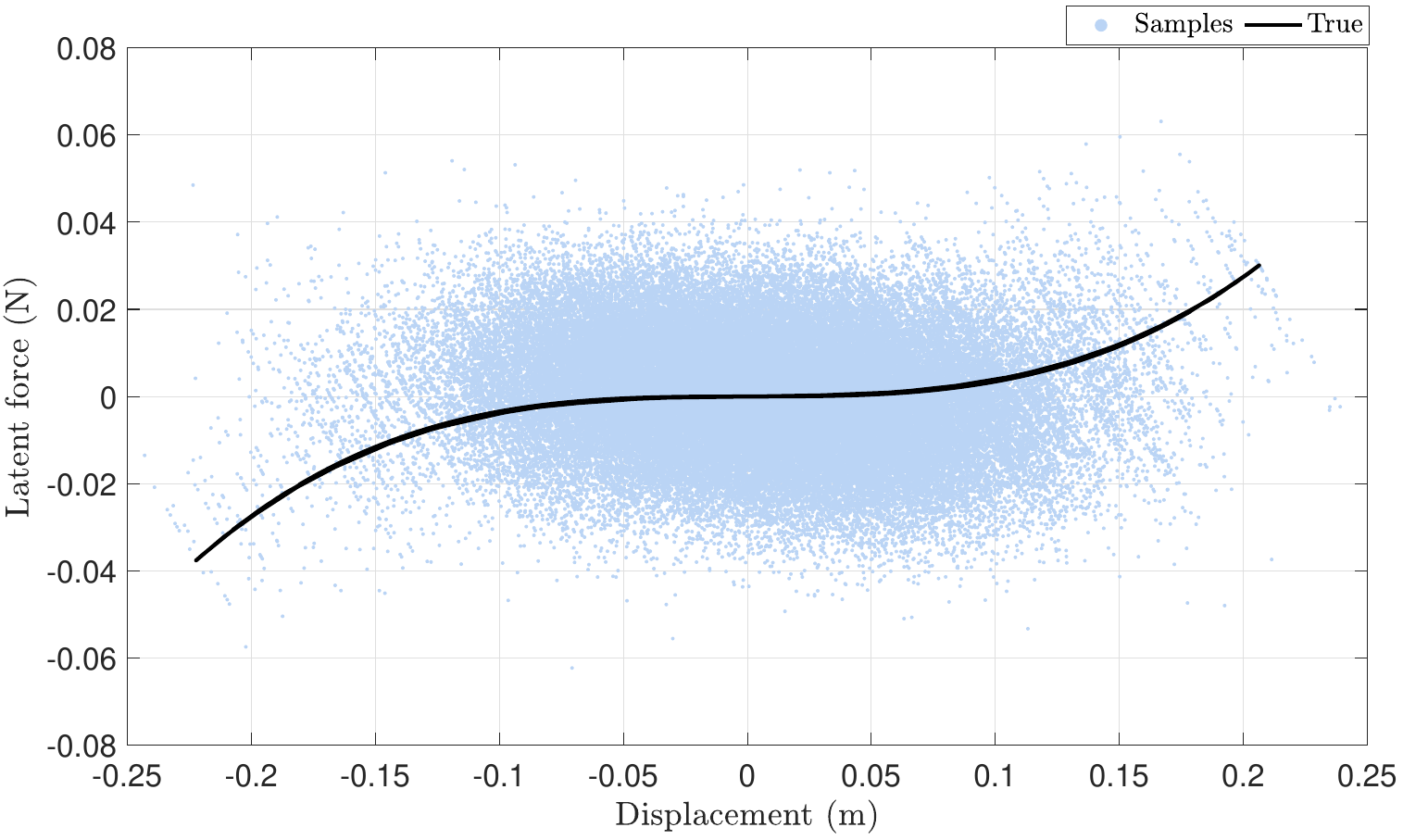}
    \caption{Posterior samples of displacement vs.\ LF from the diagnosis phase. The black line denotes the true nonlinear force $k_\text{nl} \bar{q}^3(t)$.}
    \label{fig: SilverBox_ensemble}
\end{figure}

Since the nonlinear behavior is known apriori, we omit detailed plots of state estimation. Instead, \Cref{fig: SilverBox_ensemble} shows a sample cloud from the diagnosed joint posterior of $(q(t)$, $\dot{q}(t)$, $\eta(t))$, plotted against the true nonlinear force $k_\text{nl} \bar{q}^3(t)$ (black curve). The samples reveal a strong match and confirm that the GP-based LF captures the unmodeled Duffing nonlinearity.

Following diagnosis, a BNN is trained on posterior samples to learn the probabilistic mapping $p(\eta \mid \ve{x})$, where $\ve{x} = \sbk{q\;\; \dot{q}}^T$. We reuse the same BNN architecture as in earlier examples.

\begin{figure}[H]
    \centering
    \includegraphics[width=\textwidth]{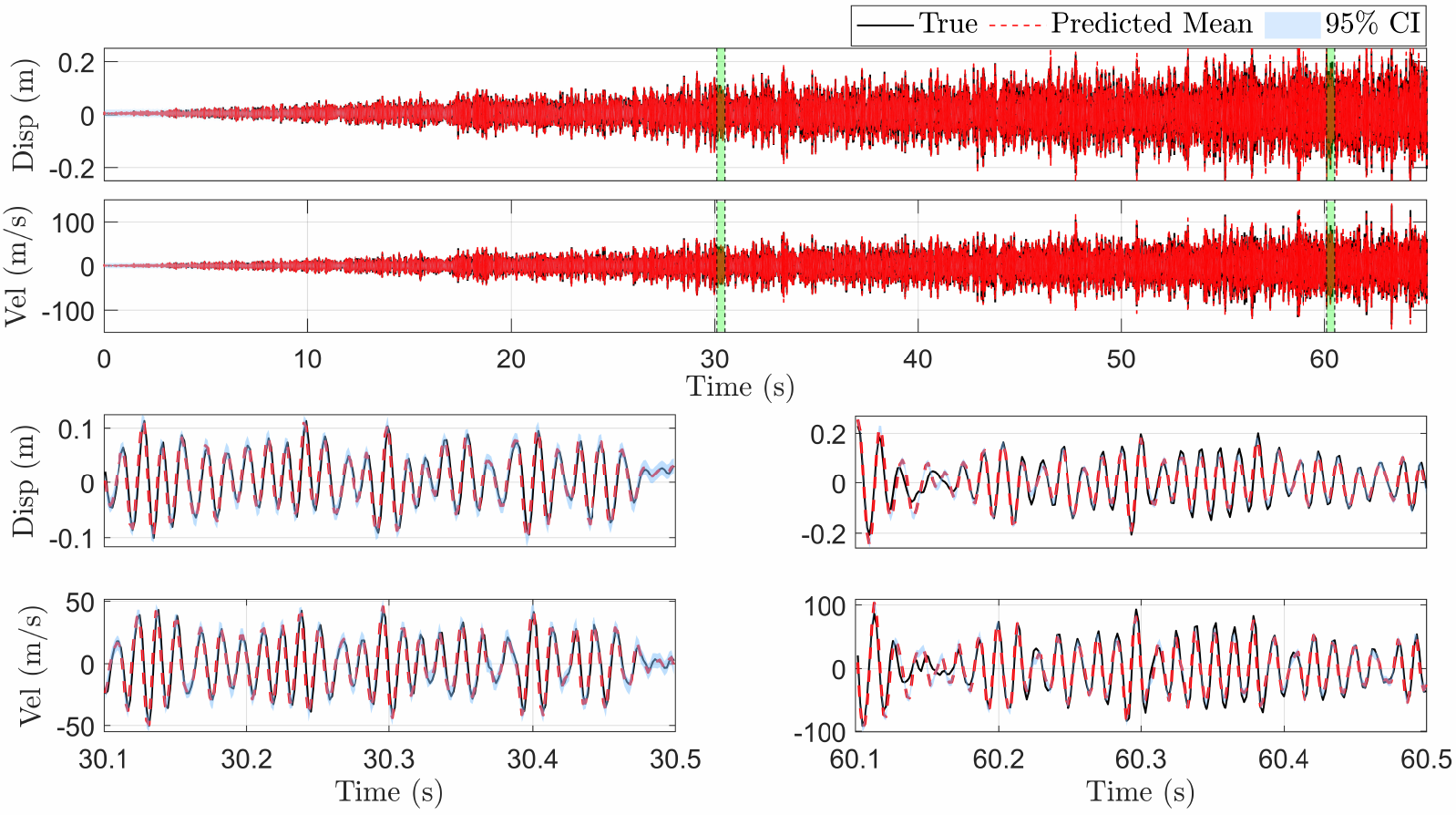}
    \caption{\textbf{Prognosis}: Predicted vs.\ true system states for the Silverbox benchmark using the GPLFM framework. Top two rows: full time histories. Bottom two rows: zoom-in views around $30\si{s}$ and $60\si{s}$. Shaded regions show $\pm 2\sigma$ confidence intervals.}
    \label{fig: SilverBox_pred}
\end{figure}

In the prognosis phase, the system is subjected to the filtered white noise (arrowhead) excitation. \Cref{fig: SilverBox_pred} presents the predicted displacement and velocity trajectories alongside true values and $95\%$ confidence intervals. Two zoomed-in regions (around $30\si{s}$ and $60\si{s}$) highlight the prediction quality under varying amplitudes. During $0$--$30\si{s}$, the prediction accuracy is excellent with NMSE less than $2\%$ for both displacement and velocity. However, as the excitation amplitude increases beyond the training regime ($\pm 0.075\si{N}$), minor degradation is observed in the predicted mean due to extrapolation. Nevertheless, even in these high-amplitude segments, NMSE remains below $6\%$, demonstrating decent generalization performance.

\subsection{Bouc-Wen Oscillator Benchmark}
To evaluate the robustness of the proposed framework under strong modeling errors, we consider a benchmark problem involving an SDOF Bouc-Wen oscillator as described in \cite{noel2016hysteretic}. This system is widely used to model hysteretic behavior in structural dynamics due to its ability to capture non-linear, memory-dependent responses under cyclic loading. Originally introduced by Bouc~\cite{bouc1967forced} and later generalized by Wen~\cite{wen1976method}, the Bouc--Wen model accounts for phenomena such as friction, plasticity, and material hysteresis.

The governing equations for the Bouc-Wen oscillator are:
\begin{subequations}
    \begin{align}
        m \ddot{\bar{q}}(t) &+ c \dot{\bar{q}}(t) + k \bar{q}(t) + b(t) = u(t), \label{eq:EoM_true_system_Bouc-Wen} \\
        \dot{b}(t) &= \alpha \dot{\bar{q}}(t) - \beta\sbk{\gamma |\bar{q}(t)| |b(t)|^{\nu - 1} b(t) + \delta \dot{\bar{q}}(t) |b(t)|^\nu }, \label{eq: Bouc-Wen_hys}
    \end{align}
\end{subequations}
where $b(t)$ denotes the nonlinear hysteretic restoring force, and the parameters $\alpha$, $\beta$, $\gamma$, $\delta$, $\nu$ control the shape of the hysteresis loop. The system parameters used are summarized in \cref{tab:boucwen_parameters}.
\begin{table}[H]
\centering
\begin{tabular}{|c|*{8}{c|}}
\hline
\textbf{Parameter}& $m$ & $c$ & $k$ & $\alpha$ & $\beta$ & $\gamma$ & $\delta$ & $\nu$ \\ \hline
\textbf{Value}  & $1\, \si{kg}$ & $10\, \si{Ns/m}$ & $5 \times 10^4\, \si{N/m}$ & $5 \times 10^4$ & $1 \times 10^3$ & 0.8 & -1.1 & 1 \\ \hline
\end{tabular}
\caption{Physical parameters of the Bouc-Wen system.}
\label{tab:boucwen_parameters}
\end{table}

In our formulation, we treat the Bouc-Wen system as the true model. As the nominal model, we consider a linear second-order oscillator that omits the hysteretic term $b(t)$. This omission introduces a significant model-form error, because the neglected term is not merely an algebraic nonlinearity but a dynamic quantity governed by its own differential equation~\eqref{eq: Bouc-Wen_hys}. We capture this unmodeled dynamics through a latent force $\eta(t)$, leading to a corrected equation of motion. The LF can be approximated as $\eta(t) \approx {b}(t)$.

\begin{figure}[H]
	\centering
	\includegraphics[width=\textwidth]{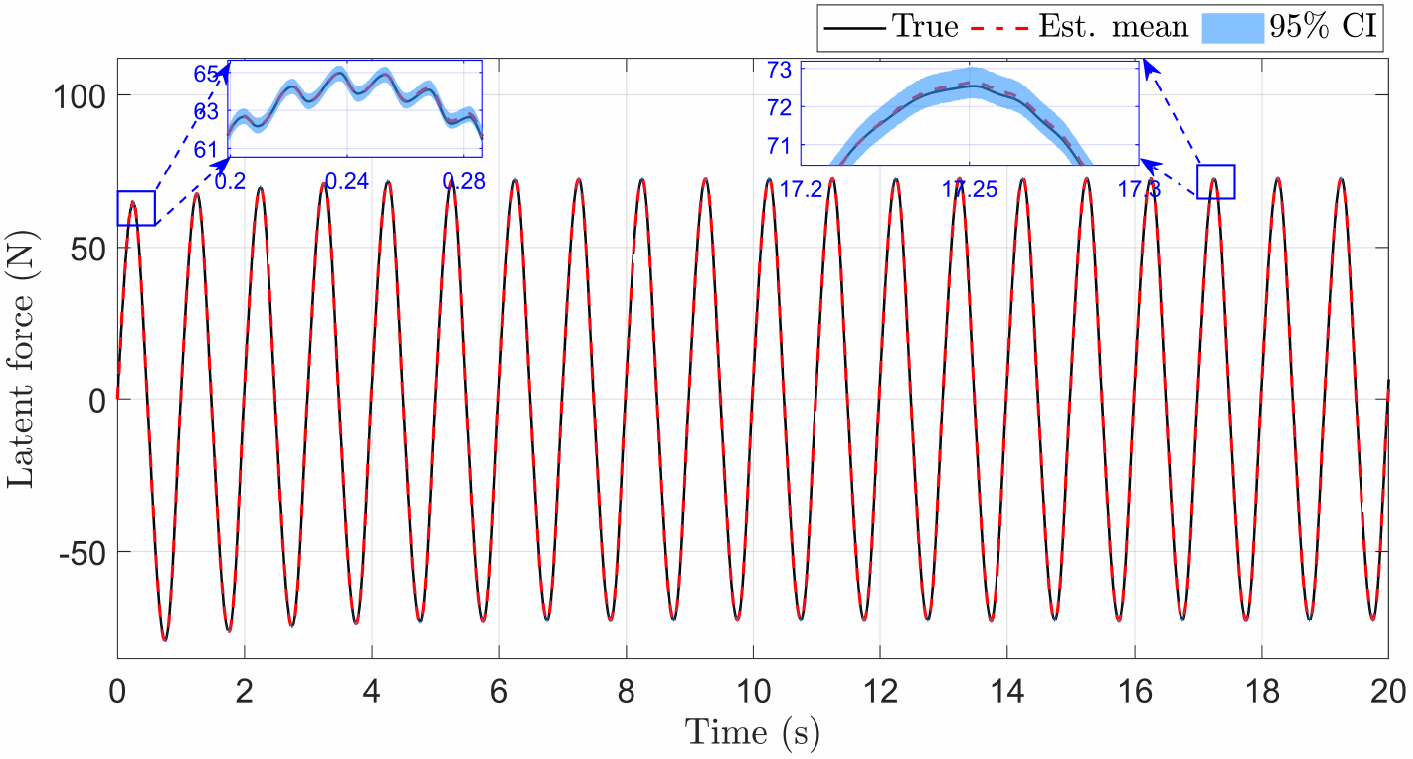}
	\caption{\textbf{Diagnosis}: Estimated Latent force versus the true hysteretic force for Bouc-Wen system with the  $\pm2\sigma$ confidence interval around the estimated mean.} 
        \label{fig: diagnosis MFE plot boucwen}
\end{figure}

\begin{figure}[H]
	\centering
	\includegraphics[width=\textwidth]{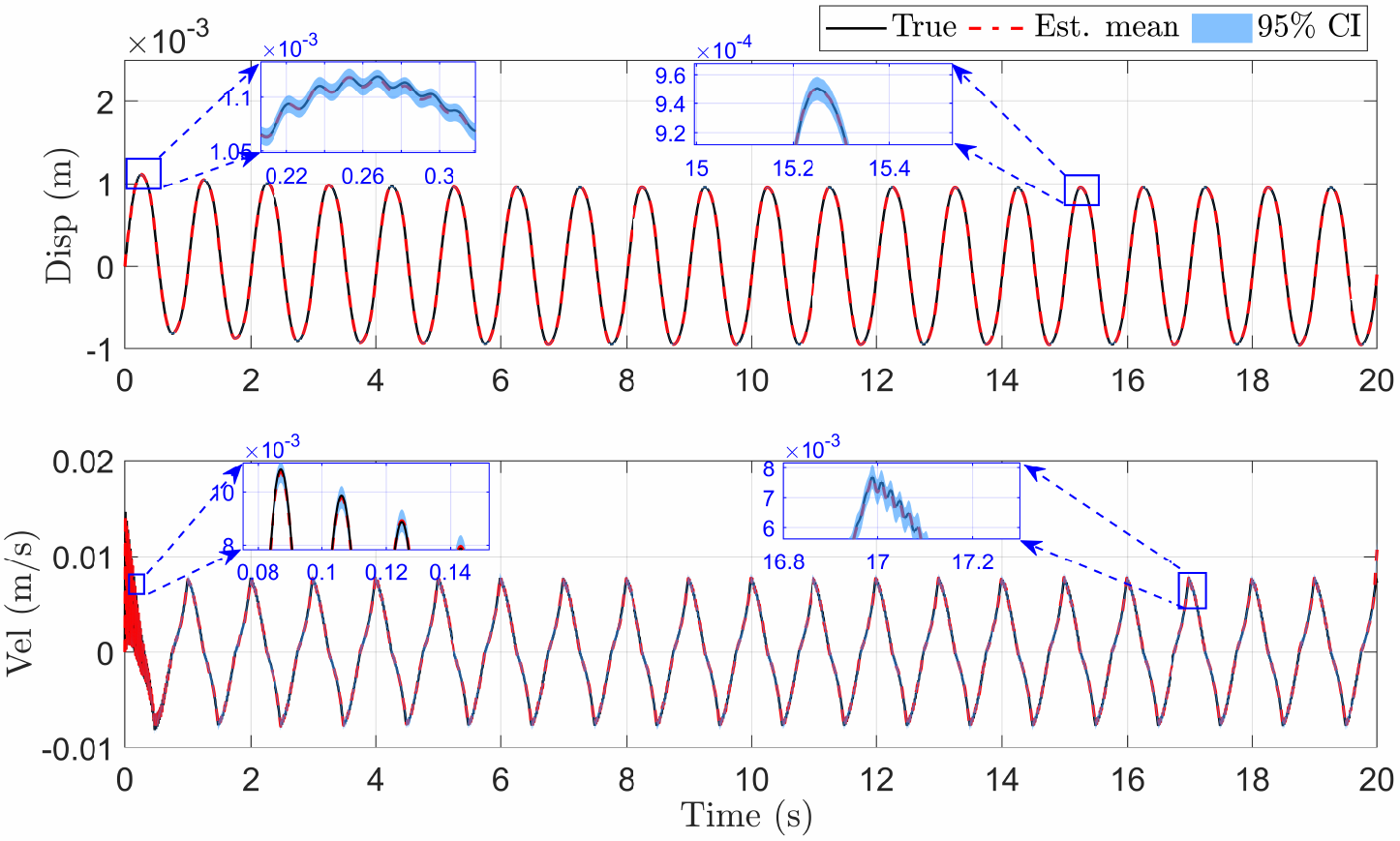}
	\caption{\textbf{Diagnosis}: Estimated displacement and velocity versus the true states, with the associated $\pm2\sigma$ confidence interval.} 
        \label{fig: diagnosis states Bouc}
\end{figure}

\paragraph{Diagnosis Phase}
We begin by diagnosing the latent dynamics using a sinusoidal input excitation of amplitude \SI{120}{N} and frequency \SI{1}{Hz}. Both noisy acceleration and displacement measurements are used for inference. The latent force $\eta(t)$ and the system states $\ve{x}(t) = [q(t) \;\; \ \dot{q}(t)]^T$ are jointly estimated using the GPLFM-based methodology described earlier. The estimated LFs and states, shown in \Cref{fig: diagnosis MFE plot boucwen,fig: diagnosis states Bouc}, match the true values closely, with narrow confidence intervals. This improved estimation performance, compared to earlier case studies, is attributed to the richer information content from the combined use of displacement and acceleration measurements. 
\begin{figure}[H]
	\centering
	\includegraphics[width=\textwidth]{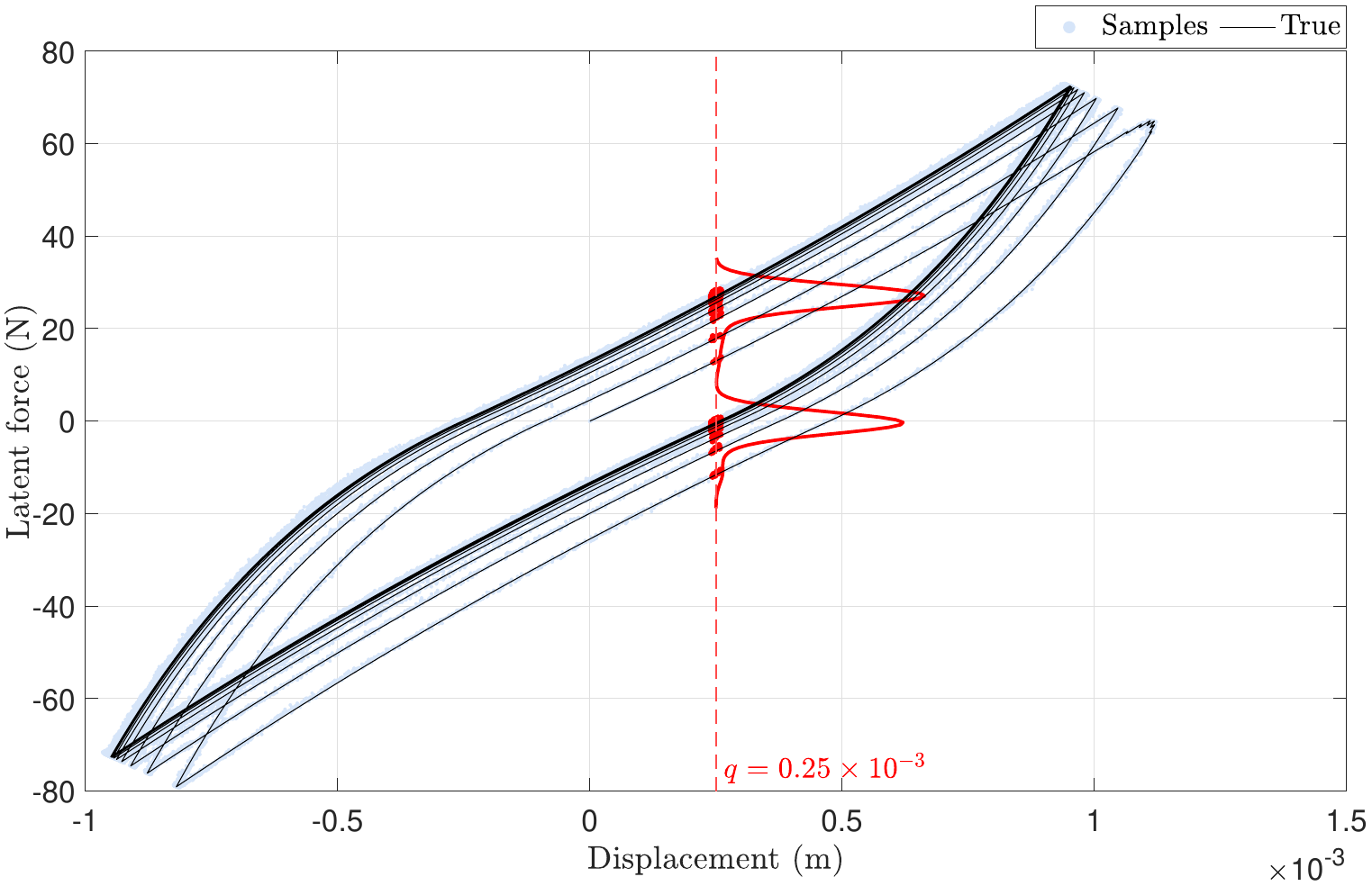}
	\caption{ Posterior samples of displacement versus LF from the diagnosis phase. The black line denotes the true nonlinear hysteretic force.} 
        \label{fig: Bouc_hyst}
\end{figure}

\paragraph{Mapping Phase}
Posterior samples of $\cbk{\ve{x}_k, \eta_k}^{N_t}_{k=1}$ obtained during diagnosis are used to learn a mapping from system states to MFE via a BNN. However, the Bouc-Wen system violates the key assumption of a static functional dependence because $ {b}(t)$ is inherently path-dependent. As shown in \Cref{fig: Bouc_hyst}, for the same displacement $q(t)$, the value of $\eta(t)$ differs depending on whether the system is loading or unloading. This yields a multimodal conditional distribution $\pr{\eta \mid \ve{x}}$, which cannot be fully captured by the unimodal Gaussian approximation used by the BNN. Nevertheless, we proceed with training a BNN to model this static mapping as a tractable approximation.

\begin{figure}[H]
	\centering
	\includegraphics[width=\textwidth]{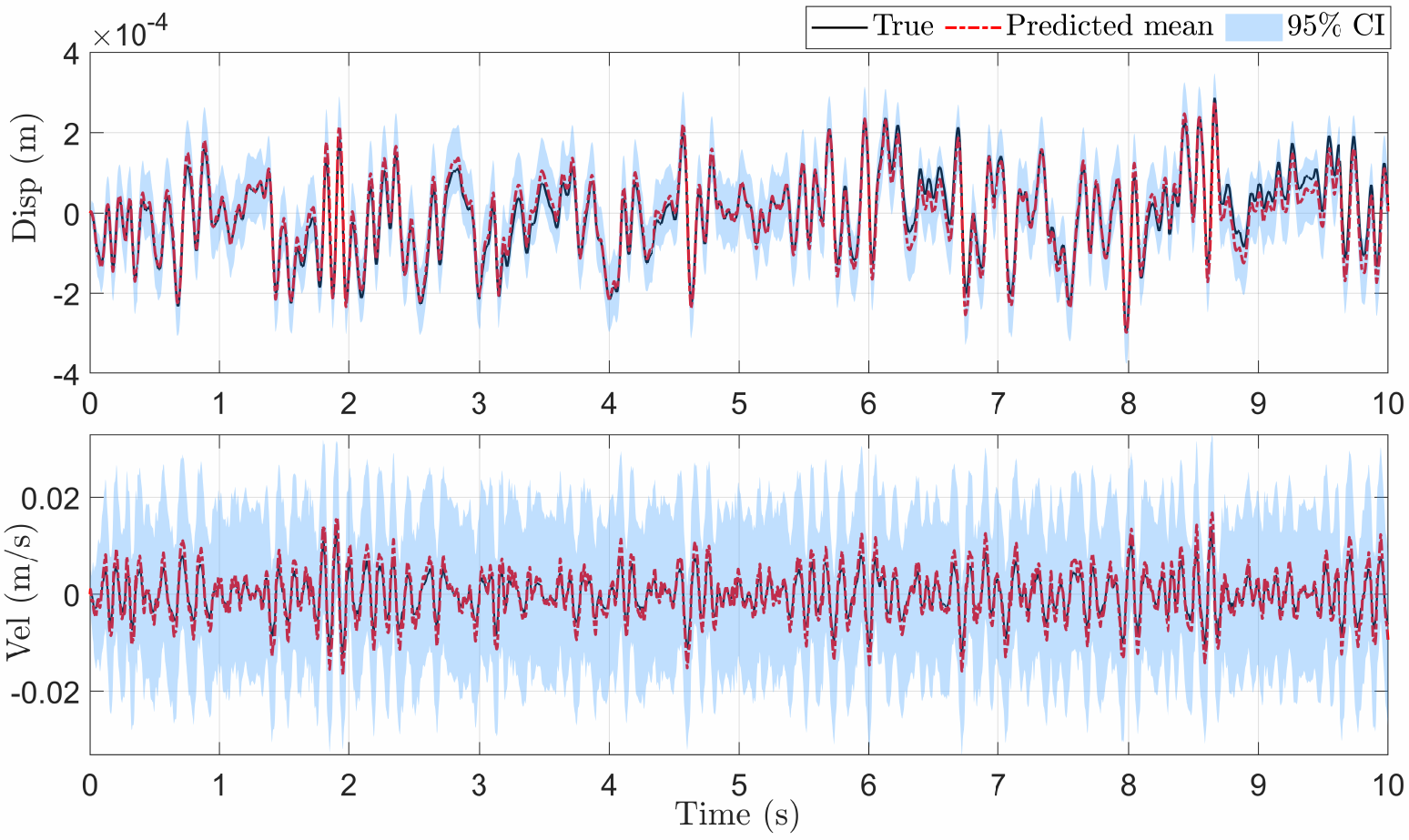}
	\caption{ \textbf{Prognosis}: Predicted system states versus true states along with the $ 2\sigma$ confidence bounds.} 
        \label{fig: Bouc_wen_pred}
\end{figure}

\paragraph{Prognosis Phase}
In the prognosis phase, the trained BNN is used to predict system states under a new excitation --- filtered white noise. The predicted states and associated uncertainty bounds are shown in \Cref{fig: Bouc_wen_pred}. While the predicted mean states exhibit some deviation from the ground truth, especially during highly nonlinear transitions, the $2\sigma$ confidence bounds reliably encompass the true trajectories. The discrepancy in the predicted means arises from the BNN’s unimodal approximation of a multimodal distribution: since the BNN cannot distinguish between loading and unloading paths, it learns an averaged response with inflated variance. This larger variance increases the pseudo-measurement noise covariance $\tilde{R}$ used in the Kalman filter, resulting in wider confidence intervals. Despite these limitations, the NMSE remains below $1\%$ for both displacement and velocity predictions, demonstrating the framework's ability to generalize even under strong model misspecification and dynamic, history-dependent nonlinearities.

\section{Discussion} \label{sec:Discussion}

\textcolor{black}{This section discusses the strengths, modeling considerations, and limitations of the proposed GPLFM-based probabilistic digital-twin framework, along with its broader implications.}

\textcolor{black}{The proposed framework offers several advantages over traditional MFE-corrected formulations. First, by embedding the BNN mapping within a Kalman filtering loop through pseudo-measurements, the approach provides stable and probabilistically consistent forward prediction. Unlike direct substitution of learned maps into governing equations, which may lead to numerical instability, this closed-loop formulation ensures numerical stability in time integration. Furthermore, the framework allows uncertainty to be consistently propagated from the diagnostic stage to prognosis, enabling calibrated confidence bounds on future predictions.}

\textcolor{black}{Although a Mat\'{e}rn-1/2 kernel is used in this study, this choice is not universally optimal. For instance, our previous work \cite{kashyap2024gaussian} showed that while the Mat\'{e}rn-3/2 kernel can yield more accurate LF estimates in certain cases, it nearly doubles the computational cost due to the larger state-space representation. Since this improvement was not consistent across all examples, the Mat\'{e}rn-1/2 kernel was adopted here as a balanced choice between accuracy and computational efficiency. More generally, the choice of kernel should be guided by the physical characteristics of the MFEs. For example, if the MFEs exhibit approximately periodic behavior, a periodic kernel may be more appropriate \cite{lourens2025MFE}. In cases where the system exhibits abrupt switching or discontinuities, switching Gaussian process models can be employed to capture regime changes effectively \cite{Alice_switchingGPLFM}. Similarly, if partial physics knowledge about the MFE is available, one may design physics-informed kernels to embed this information within the GP prior \cite{Cross_physicsGP}. Overall, the proposed methodology is flexible and can incorporate any covariance kernel that admits a stable SDE representation. The kernel should thus be selected based on the expected characteristics of the underlying MFEs.}

\textcolor{black}{In this work, the continuous-time process noise covariance $\matr{Q}_c$ is modeled as block-diagonal, reflecting the assumption of mutually independent GP-based latent forces associated with each degree of freedom. If the latent forces are functions of the system states, a full covariance matrix is required to capture the cross-correlations between the states and latent forces. However, assuming independence substantially reduces the number of parameters to be optimized, while still providing reasonable performance. Because the framework requires batch optimization for GP hyperparameter, maintaining a compact parameter space is essential for computational tractability. Incorporating correlated GP priors (resulting in a non-block-diagonal $\matr{Q}_c$) can more accurately represent state-dependent model errors but would significantly increase computational cost and complexity. Extending the current formulation to include such correlations remains an interesting direction for future research.}

\textcolor{black}{While the framework has several advantages, certain limitations and challenges remain. The framework assumes a static conditional map $\pr{\ve{\eta} \mid \ve{x}}$ between MFEs and states, learned via a BNN. This assumption breaks down in systems with history-dependent dynamics. For example, in the Bouc–Wen oscillator, the hysteretic force depends not only on the current state but also on its loading history. As shown in our results, this leads to multimodal $\ve{\eta}$ distributions for a fixed $\ve{x}$, which cannot be captured by a unimodal static BNN map. The result is inflated uncertainty. Furthermore, prognosis quality is directly tied to the accuracy of the BNN mapping and also the diagnosis inference. As observed in the Silverbox benchmark, extrapolation to test scenarios outside the training domain (i.e., unseen excitation amplitudes or state ranges) led to performance degradation. This underscores the need for training data that spans a wide and representative range of state-space regions. Additionally, since GP hyperparameters are optimized in batch mode, the current implementation is primarily suited for offline applications, although online extensions are possible.}

\textcolor{black}{In many cases, measurements are available only at a subset of the system’s DOFs, depending on where sensors are installed. As shown in our previous work \cite{kashyap2024gaussian}, when sensors are placed sparsely, the GPLFM-based diagnosis framework may face challenges in correctly identifying and localizing the MFEs. Hence, sensor placement must be designed carefully to ensure reliable estimation. While the present study focuses on demonstrating the complete end-to-end probabilistic digital-twin framework, the findings from our earlier analysis emphasize the importance of systematic sensor placement optimization.}

\textcolor{black}{Finally, the proposed framework is particularly relevant to the development of digital twins, where reliable prediction of future states under evolving operating conditions is critical. Digital twins aim to tightly couple physics-based models with real-time sensor data to enable continuous monitoring, diagnostics, and control. However, a major bottleneck in realizing robust digital twins is the presence of MFEs arising from unmodeled dynamics or simplified assumptions. By integrating MFE diagnosis, probabilistic mapping, and uncertainty-aware prognosis into a unified pipeline, our framework offers an approach to enhance predictive fidelity. Furthermore, the ability to propagate uncertainty from sensor data to future predictions is vital for decision-making tasks within digital twins. While our current implementation operates offline, extensions to online learning and inference would make the approach suitable for real-time digital twin applications.}

\section{Conclusion and Future work} \label{sec:conclusion}

This study presented a novel end-to-end probabilistic framework for accurate state prediction in dynamical systems governed by misspecified physics. The approach is structured into three stages: (i)~\textit{Diagnosis}, which uses GPLFM-based Kalman filtering and smoothing to estimate the system states and MFE jointly, (ii)~\textit{Mapping}, which learns a probabilistic relationship between states and MFEs using a BNN, and (iii)~\textit{Prognosis}, which propagates this mapping within a Kalman filtering and smoothing framework to predict future states under new external force excitations without further measurements.

Unlike conventional MFE-corrected formulations that directly embed a learned mapping into the governing equations and solve them using numerical integration --- often resulting in unstable trajectories --- the proposed framework avoids this pitfall by treating the learned MFE-state relationship as a pseudo-measurement. This enables stable probabilistic inference during prediction, even when the true dynamics are nonlinear and unknown.

The effectiveness of the framework was demonstrated on both simulated and experimental benchmark systems, including cases with local nonlinearities, cubic stiffness, and dynamic hysteresis. Results showed that the proposed method provides accurate state predictions with well-calibrated uncertainty estimates.

Overall, this work bridges model-based and data-driven paradigms within a probabilistic framework, offering a flexible, interpretable, and computationally robust methodology for response prediction in misspecified dynamical systems. Future work could extend this framework to online learning settings, incorporate history-aware mappings, and explore nonlinear filtering methods for nominally nonlinear systems.
\section*{Acknowledgements}
\noindent \textbf{R. Nayek} acknowledges the financial support received from the Science and Engineering Research Board (SERB) via grant number SRG/2022/001410, from Ministry of Ports, Shipping and Waterways via project no. ST-14011/74/2022-MT (356529) and the matching grant received from IIT Delhi. 

\appendix

\section{Kalman filter and RTS smoother} \label{appendix:Kfs}
The discretized version of the process equation in \cref{eq:augSSM_matrix_form}, along with the measurement equation in \cref{eq:MeasaugSSM_matrix_form}, for a sampling interval $\Delta t$, is expressed as:
\begin{align} \label{eq:disc_SSM}
\ve{z}_k &= \matr{F}_d \ve{z}_{k-1} + \matr{B}^a_{ud} \ve{u}_{k-1} + \matr{B}^a_{gd} {\ddot{u}}_{g,k-1}  +  \ve{w}_{k-1}
\\ \ve{y}_k &= \matr{H}^a\ve{z}_k + \matr{J}_{u,k} +\ve{v}_k
\end{align}
where $\matr{F}_d =\exp{(\matr{F}_c\Delta{t})}$, $\matr{B}^a_{ud} = {\matr{F}_c}^{-1}\sbk{\matr{F}_d -\matr{I} }\matr{B}^a_{uc}$ and, $\matr{B}^a_{gd} = {\matr{F}_c}^{-1}\sbk{\matr{F}_d -\matr{I} }\matr{B}^a_{gc}$. $\ve{z}_0 \sim \mathcal{N}({\ve{m}}_{0|0}, \matr{P}_{0|0})$ represents the initial state distribution, $\ve{w}_{k-1} \sim \mathcal{N}(\zeros, \matr{Q}_d)$ and $\ve{v}_{k} \sim \mathcal{N}(\zeros, \matr{R})$ are the process noise and the measurement noise distribution respectively. Here $\matr{Q}_d = \int_{0}^{\Delta{t}} \matr{\Psi}{(\Delta{t} - \tau)} \matr{Q}_c \matr{\Psi}{(\Delta{t} - \tau)}^T \,d\tau\,$, with $\matr{\Psi}=\exp{(\matr{F}_c\tau)}$ representing the matrix exponential of the state-transition matrix. This integral can be solved using matrix
fraction decomposition as implemented in \cite{sarkka2006recursive}.   
Now, for this system we use Kalman filter \cite{kalman1960new} to get estimates of state mean and covariance using the following steps.  
\begin{align}
{\ve{m}}_{k|k-1} &= \matr{F}_d  {\ve{m}}_{k-1|k-1} + \matr{B}^a_{ud} \ve{u}_{k-1} + \matr{B}^a_{gd} \ddot{u}_{k-1}   \\
{\matr{P}}_{k|k-1} &= \matr{F}_d \matr{P}_{k-1|k-1} \matr{F}_d^{{T}} + \matr{Q}_d \\
\ve{e}_k &= \ve{y}_k - ({\matr{H}^a {\ve{m}}_{k|k-1} + \matr{J_u} \ve{u}_k})\label{eq:innov}\\
\matr{S}_k & = \matr{H}^a {\matr{P}}_{k|k-1} \brc{\matr{H}^a}^{{T}} + \matr{R}\label{eq:innov cov}\\
\matr{K}_k &= {\matr{P}}_{k|k-1} \brc{\matr{H}^a}^{{T}} \matr{S}_k^{-1} \\
{\ve{m}}_{k|k} &= {\ve{m}}_{k|k-1} + \matr{K}_k \ve{e}_k \\
\matr{P}_{k|k} &= {\matr{P}}_{k|k-1} - \matr{K}_k \matr{S}_k \matr{K}^{{T}}_k
\end{align}
for $k = 1,\dots,N_t$. Here ${\ve{m}}_{k|k-1}$ and $\ve{m}_{k|k}$ represent 
the $k^{\text{th}}$ predicted and filtered state estimate, respectively, and, ${\matr{P}}_{k|k-1}$ 
and $\matr{P}_{k|k}$ denote the $k^{\text{th}}$ predicted and filtered state error covariance 
matrices respectively. The recursion in Kalman filter is started from the initial 
state ${\ve{m}}_{0|0}$ and covariance $\matr{P}_{0|0}$. Following filtering step, 
the (fixed interval) smoothing step is given by the RTS smoother \cite{rauch1965maximum}:
\begin{align}
\matr{N}_k &= \matr{P}_{k|k} \matr{F}_d^{{T}} \left(\matr{P}_{k+1|k}\right)^{-1}\\
{\ve{m}}_{k|N} &= {\ve{m}}_{k|k} + \matr{N}_k \left({\ve{m}}_{k+1|N} - 
{\ve{m}}_{k+1|k} \right) \\
\matr{P}_{k|N} &= \matr{P}_{k|k} +  \matr{N}_k \left(\matr{P}_{k+1|N} - \matr{P}_{k+1|k} \right) 
\matr{N}_k^T 
\end{align}
for $k = N_t-1,\ldots,1$.
and the recursion is started from the last time step with ${\ve{m}}_{k|N}= 
{\ve{m}}_{N|N}$ and $\matr{P}_{N|N}=\matr{P}_N$
\section{Kullback-Liebler Divergence between two Gaussians}\label{appendix: KLD}
Let $\ve{q}\sim\mathcal{N}(\ve{\mu}^q, \matr{\Sigma}^q)$ and $\ve{p}\sim\mathcal{N}(\ve{\mu}^p, \matr{\Sigma}^p)$ be two multivariate Gaussian distribution of 
$\text{J}$-dimension. The Kullback-Leibler (KL) divergence between  distributions $\ve{q}$ and $\ve{p}$ is given as
\begin{align}
\mathrm{KL}(\ve{q}(\ve{\varphi}) \mid\mid \ve{p}(\ve{\varphi})) =
\int_{\ve{\varphi}} \ve{q}(\ve{\varphi}) 
\log \frac{\ve{q}(\ve{\varphi})}{\ve{p}(\ve{\varphi})} 
\; d\ve{\varphi}
\end{align} 
This can be expressed using the expectation operator as 
\begin{align}
   \mathrm{KL}(\ve{q}(\ve{\varphi}) \mid\mid \ve{p}{(\ve{\varphi})}) &= \mathbb{E}_q \sbk{ \log \ve{q}(\ve{\varphi}) - \log \ve{p}(\ve{\varphi}) }
\end{align}

Hence,
\begin{align}
\mathrm{KL}(\ve{q}(\ve{\varphi})\mid\mid \ve{p}(\ve{\varphi})) 
&= \mathbb{E}_{q} \sbk{ \log \ve{q}(\ve{\varphi}) - \log \ve{p}(\ve{\varphi}) } \\
&= \mathbb{E}_{q} \biggl[
    \frac{1}{2} \log \frac{|\matr{\Sigma}^p|}{|\matr{\Sigma}^q|}
    - \frac{1}{2} 
        (\ve{\varphi} - \ve{\mu}^q)^\top 
        (\matr{\Sigma}^q)^{-1} 
        (\ve{\varphi} - \ve{\mu}^q) \label{eq: KLD_trace_equivalance}\\
& \qquad\quad 
    + \frac{1}{2} 
        (\ve{\varphi} - \ve{\mu}^p)^\top 
        (\matr{\Sigma}^p)^{-1} 
        (\ve{\varphi} - \ve{\mu}^p)
\biggr] \\
&= \frac{1}{2} \mathbb{E}_{q}\sbk{\log 
    \frac{|\matr{\Sigma}^p|}{|\matr{\Sigma}^q|}} 
    - \underbrace{\frac{1}{2} 
        \mathbb{E}_q 
        \Bigl[
            (\ve{\varphi} - \ve{\mu}^q)^\top 
            (\matr{\Sigma}^q)^{-1} 
            (\ve{\varphi} - \ve{\mu}^q)
        \Bigr]}_\text{A}\\
& \qquad\quad
    + \underbrace{\frac{1}{2} 
        \mathbb{E}_q 
        \Bigl[
            (\ve{\varphi} - \ve{\mu}^p)^\top 
            (\matr{\Sigma}^p)^{-1} 
            (\ve{\varphi} - \ve{\mu}^p)
        \Bigr]}_\text{B}
\end{align}

Let us first evaluate term A. Note that the quadratic term $(\ve{\varphi} - \ve{\mu}^q)^\top 
        (\matr{\Sigma}^q)^{-1} 
        (\ve{\varphi} - \ve{\mu}^q)$ can be equivalently expressed using the trace operator $tr\cbk{.}$ (using Eq. 16 in \cite{cookBook}) as \\$tr\cbk{(\ve{\varphi} - \ve{\mu}^q)(\ve{\varphi} - \ve{\mu}^q)^\top 
        (\matr{\Sigma}^q)^{-1} 
        }$.
Moving the expectation operator inside the trace gives
\begin{align}
    \frac{1}{2}tr\cbk{\mathbb{E}_q\Bigl[(\ve{\varphi} - \ve{\mu}^q)(\ve{\varphi} - \ve{\mu}^q)^\top 
        (\matr{\Sigma}^q)^{-1} 
        \Bigr]}
\end{align}
We know that
\begin{align}
    \mathbb{E}_q\Bigl[(\ve{\varphi} - \ve{\mu}^q)(\ve{\varphi} - \ve{\mu}^q)^\top 
        \Bigr] &= \matr{\Sigma}^q\\
        \end{align}
Hence,
\begin{align}\frac{1}{2}tr\cbk{\mathbb{E}_q\Bigl[(\ve{\varphi} - \ve{\mu}^q)(\ve{\varphi} - \ve{\mu}^q)^\top 
        (\matr{\Sigma}^q)^{-1} 
        \Bigr]}&=\frac{1}{2} tr\cbk{ \matr{\Sigma}^q  (\matr{\Sigma}^q)^{-1} }\\
        &=\frac{1}{2}tr\cbk{\matr{I}_\text{J}}\\
        &=\frac{\text{J}}{2}
\end{align} 
Now, let us evaluate term B. Using Eq. 380 from \cite{cookBook}, we obtain
\begin{align}
    \frac{1}{2} 
        \mathbb{E}_q 
        \Bigl[
            (\ve{\varphi} - \ve{\mu}^p)^\top 
            (\matr{\Sigma}^p)^{-1} 
            (\ve{\varphi} - \ve{\mu}^p)
        \Bigr] = \frac{1}{2} \brc{\ve{\mu}^q-\ve{\mu}^p}^\top\brc{\matr{\Sigma}^q}^{-1}\brc{\ve{\mu}^q-\ve{\mu}^p} + tr\cbk{\brc{\matr{\Sigma}^q}^{-1}{\matr{\Sigma}^q}}
\end{align}
Combining these results, we finally obtain
\begin{align}\label{eq: KLD_MVD}
    \mathrm{KL}(\ve{q}(\ve{\varphi})\mid\mid \ve{p}(\ve{\varphi})) =  \frac{1}{2} \sbk{\log 
    \frac{|\matr{\Sigma}^p|}{|\matr{\Sigma}^q|} - \text{J} + \brc{\ve{\mu}^q-\ve{\mu}^p}^\top\brc{\matr{\Sigma}^q}^{-1}\brc{\ve{\mu}^q-\ve{\mu}^p} + tr\cbk{\brc{\matr{\Sigma}^q}^{-1}{\matr{\Sigma}^q}}}
\end{align}

In the BNN formulation the $\mathrm{KL}$ is taken between two fully factorized Gaussian distribution hence \cref{eq: KLD_MVD} reduces to
\begin{align}
\mathrm{KL} \brc{ 
\mathcal{N}
\brc{ 
(\mu_1^q, \dots, \mu_\text{J}^q)^\top, \,
\texttt{diag}((\sigma_1^q)^2, \dots, (\sigma_\text{J}^q)^2)
} \mid\mid \mathcal{N}(\ve{0}, \matr{I})
}
= 
\sum_{j=1}^\text{J} 
\sbk{-\log \sigma_j^q
+
\frac{(\sigma_j^q)^2 + (\mu_j^q)^2}{2}-\frac{1}{2}} 
\end{align}

\section{\textcolor{black}{GP hyperparameters optimization}}\label{appendix: GP hyperparameter Opti}
\textcolor{black}{In the methodology adopted in this study, each $i^\text{th}$ DOF is associated with an independent GP in the form of an LF $\eta_i$ to capture the MFEs. The hyperparameters of these GPs, $\theta_i = \{\alpha_i, \ell_i\}$, are to be optimized. For this purpose, they are treated as random variables, and their posterior distribution is obtained from the product of the data likelihood and the prior distribution. Given the structural parameters $\phi$, input force $\ve{u}_{1:N_t}$, and measurements $\ve{y}_{1:N_t}$, the posterior distribution can be written as
\begin{align}
    \pr{\vtheta \mid \ve{y}_{1:N_t},\ve{\phi},\ve{u}_{1:N_t}} 
    \propto \pr{\ve{y}_{1:N_t} \mid \vtheta,\ve{\phi},\ve{u}_{1:N_t}} \, \pr{\vtheta}.
\end{align}}

\textcolor{black}{For SSMs, the data likelihood is obtained by marginalizing out the states $\ve{z}_{1:N_t}$; thus, $\pr{\ve{y}_{1:N_t} \mid \vtheta,\ve{\phi},\ve{u}_{1:N_t}}$ is also referred to as the marginal likelihood, while $\pr{\vtheta}$ denotes the prior over the GP hyperparameters. As explained in the main text, to promote sparsity, we impose independent Student's-$t$ prior distributions on $\alpha_i$ and $\ell_i$. The Student's-$t$ distribution, denoted by $\mathcal{T}(\mu_t, v_t, \nu)$, is parameterized by its mean $\mu_t$, variance $v_t$, and statistical degrees of freedom $\nu$. The parameter $\nu$ controls the heaviness of the distribution’s tails, thereby enforcing sparsity. In particular, $\nu=1$ yields the heaviest tails, which is the choice adopted in this work. Hence,
\begin{align}
    \log \pr{\vtheta} = \sum_{j=1}^n \left[ 
        \log \mathcal{T}(\alpha_j; 0, 1, 1) 
        + \log \mathcal{T}(\ell_j; 100, 10, 1)
    \right],
\end{align}
where independent $t$-priors are assigned to $\alpha_j$ with zero mean and unit variance, while $\ell_j$ are assigned $t$-priors with mean $100$ and variance $10$.}

\textcolor{black}{The marginal likelihood is factorized recursively as
\begin{align}
    \pr{\ve{y}_{1:N_t} \mid \vtheta,\ve{\phi},\ve{u}_{1:N_t}}
    = \prod_{k=1}^{N_t} \pr{\ve{y}_k \mid \ve{y}_{1:k-1}, \vtheta,\ve{\phi}, u_k}.
\end{align}
For linear Gaussian SSMs, the right-hand side can be evaluated in closed form using the Kalman filter's innovations $\brc{\ve{e}_k}$ and innovation covariances $\brc{\matr{S}_k}$ as seen in  \cref{eq:innov,eq:innov cov}. For more details refer \cite{sarkka2013bayesian}):
\begin{align}
    \log \pr{\ve{y}_{1:N_t} \mid \vtheta,\ve{\phi},\ve{u}_{1:N_t}}
    = \sum_{k=1}^{N_t} \log \pr{\ve{y}_k \mid \ve{y}_{1:k-1}, \vtheta,\ve{\phi}, u_k}
    = - \sum_{k=1}^{N_t} \left[ \log \det \matr{S}_k 
    + \ve{e}_k^T \matr{S}_k^{-1} \ve{e}_k \right].
\end{align}}

\textcolor{black}{Finally, the MAP estimate of the hyperparameters is obtained by minimizing the negative log-posterior:
\begin{align}
    \hat{\vtheta}_{\text{MAP}}
    &= \underset{\vtheta}{\arg\min} 
    \left\{ - \log \pr{\ve{y}_{1:N_t} \mid \vtheta, \ve{\phi}, \ve{u}_{1:N_t}}
    - \log \pr{\vtheta} \right\}.
\end{align}
This optimization is performed in the log-space of $\vtheta$, thereby constraining the hyperparameters to remain positive.}

\section{\textcolor{black}{Computational Cost Analysis}}\label{appendix: comp_cost}
\textcolor{black}{To assess the computational cost of the proposed GPLFM framework, we consider four shear-storey systems with increasing dimensionality: SDOF, 3DOF, 5DOF, and 7DOF. In all cases, the true source of MFE is a Duffing-type spring acting between the ground and the first mass. Since the exact location of the MFE is not known \emph{apriori}, latent forces are introduced at all DOF.}

\textcolor{black}{Two metrics are used to quantify computational effort: (i) the number of floating-point operations (FLOPs), measured using the \texttt{Lightspeed} toolbox \cite{minka2011lightspeed}, and (ii) the wall-clock time, measured via MATLAB’s \cite{MATLAB:R2023b_u9} \texttt{tic}/\texttt{toc} functions. Each optimization iteration corresponds to one objective function evaluation (i.e., one complete forward filtering pass), so both metrics are reported \emph{per iteration}. This allows for fair comparison independent of optimizer settings or convergence criteria. The total optimization cost can be obtained by multiplying the per-iteration cost by the number of iterations required for convergence.}
\begin{figure}[H]
	\centering
	\includegraphics[width=\textwidth]{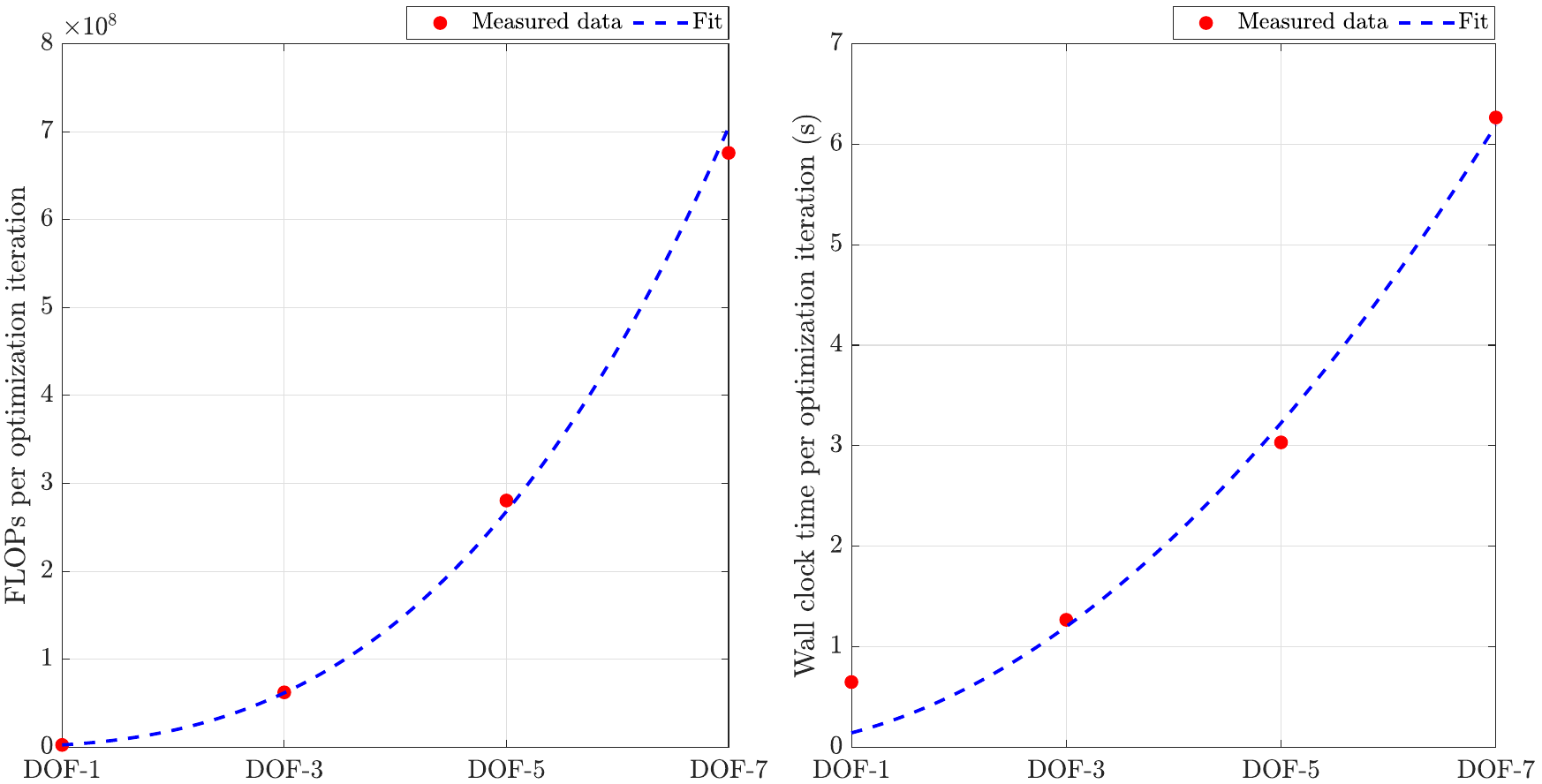}
	\caption{\textcolor{black}{Scaling of computational cost with the number of DOFs for the GPLFM framework. \textbf{Left:} FLOPs per optimization iteration and the corresponding power-law fit. \textbf{Right:} Wall-clock time per optimization iteration and the corresponding power-law fit. Reporting per-iteration metrics allows comparison independent of optimizer settings.}}
        \label{fig:comp_cost_vs_dof}
\end{figure}
\textcolor{black}{\cref{fig:comp_cost_vs_dof} shows how both metrics vary with the number of DOFs. As expected, the computational cost increases nonlinearly with system size, with the FLOP count growing approximately cubic and the wall-clock time increasing nearly quadratically. The near-cubic scaling reflects the matrix operations inherent to Kalman filtering, while the quadratic (sub-cubic) growth in runtime likely benefits from efficient vectorization and internal numerical optimizations in MATLAB.}

\textcolor{black}{Overall, the analysis shows that while the GPLFM framework is computationally more involved than a standard linear filter, the additional cost due to GP augmentation and hyperparameter learning scales predictably and remains tractable for systems of moderate dimension. For large-scale systems, reduced-order formulations or modal projections can further mitigate computational cost.}

\bibliographystyle{elsarticle-num-names} 
\bibliography{ref.bib}
\end{document}